\documentclass{article} %
\usepackage{colm2024_conference}

\usepackage{booktabs}
\usepackage{graphicx}
\usepackage{enumitem}
\usepackage{wrapfig}
\usepackage{algorithm}
\usepackage{algpseudocode}
\usepackage{wrapfig}
\usepackage{float}
\usepackage{microtype}
\usepackage{amsmath}
\usepackage{amssymb}
\usepackage{colortbl}
\usepackage[utf8]{inputenc}
\usepackage{caption}
\usepackage{subcaption}
\usepackage{xcolor}
\usepackage{setspace}
\usepackage{url}
\usepackage{multirow}
\usepackage{colortbl}
\usepackage{tabularx}
\usepackage{blindtext}
\usepackage{pgfplots}
\pgfplotsset{compat=1.18}
\usepackage{tikz}
\usetikzlibrary{er, positioning, bayesnet}
\usepackage{makecell}
\usepackage{tipa}
\usepackage{siunitx}
\usepackage{nicefrac}
\usepackage{tocloft}
\usepackage{listings}
\usepackage[raster, skins]{tcolorbox} %
\usepackage{xltabular}
\usepackage{adjustbox}
\usepackage{xurl}
\usepackage{longtable}  

\usepackage{amsmath,amsfonts,bm}

\def\eqref#1{equation~\ref{#1}}

\def\1{\bm{1}}

\DeclareMathAlphabet{\mathsfit}{\encodingdefault}{\sfdefault}{m}{sl}
\SetMathAlphabet{\mathsfit}{bold}{\encodingdefault}{\sfdefault}{bx}{n}

\definecolor{lightgray}{rgb}{0.9,0.9,0.9}

\usepackage{afterpage}
\usepackage{cuted}
\usepackage{stfloats}
\usepackage{ulem}           
\usepackage{soul}           
\usepackage{tablefootnote}  
\usepackage{comment}        

\usepackage{pifont}         
\usepackage{wasysym}

\usepackage{longtable}
\usepackage{array}
\usepackage{caption}  
\usepackage{ragged2e} 

\definecolor{lightblue}{RGB}{173,216,230}
\definecolor{lightgreen}{RGB}{144,238,144}
\definecolor{lightpink}{RGB}{255, 228, 225}
\definecolor{lightred}{RGB}{255,182,193}
\definecolor{lightyellow}{RGB}{255,255,224}
\definecolor{lightpurple}{RGB}{221,160,221}
\definecolor{lightgray}{RGB}{211,211,211}
\definecolor{lightorange}{RGB}{255,218,185}
\definecolor{lightpeach}{rgb}{1.0, 0.882, 0.788}
\definecolor{lightcyan}{rgb}{0.8196, 0.9725, 0.9804}
\definecolor{sh_blue}{rgb}{0,0.60,0.93}
\definecolor{sh_red}{rgb}{0.8627, 0.3098, 0.3176}
\definecolor{highlight}{RGB}{255,255,0}
\definecolor{warning}{RGB}{255,99,71}
\definecolor{success}{RGB}{50,205,50}
\definecolor{info}{RGB}{30,144,255}
\definecolor{top1}{RGB}{255,179,179}
\definecolor{top2}{RGB}{255,217,179}
\definecolor{top3}{RGB}{255,255,179}
\definecolor{textblue}{RGB}{94,159,220} 
\definecolor{textgreen}{RGB}{59,125,35} 
\definecolor{textorange}{RGB}{192,80,21} 
\definecolor{tagred}{RGB}{196,15,15} 
\definecolor{tagblue}{RGB}{33,95,154} 
\definecolor{teaserblue}{RGB}{33,95,154} 
\definecolor{teasergree}{RGB}{57,158,163} 
\definecolor{teaserpurpe}{RGB}{105,111,173} 

\newcommand{\cmark}{\color{teasergree}\ding{51}}
\newcommand{\xmark}{\color{tagred}\ding{55}}

\newcommand{\param}[1]{\textcolor{teaserblue}{#1}}  

\definecolor{primary}{RGB}{70,130,180}
\definecolor{secondary}{RGB}{119,136,153}
\definecolor{accent}{RGB}{255,140,0}

\definecolor{customblue}{HTML}{E7EFFA}
\definecolor{custompink}{HTML}{F7E1ED}

\renewcommand{\arraystretch}{1.25}

\makeatletter
\DeclareRobustCommand\onedot{\futurelet\@let@token\@onedot}
\def\@onedot{\ifx\@let@token.\else.\null\fi\xspace}

\makeatother

\title{
JarvisEvo: Towards a Self-Evolving Photo Editing Agent\\ with Synergistic Editor-Evaluator Optimization}

\author{ \bf Tencent Hunyuan \quad Xiamen University}

\begin{document}

\maketitle

\begin{abstract}

Agent-based editing models have substantially advanced interactive experiences, processing quality, and creative flexibility. However, two critical challenges persist: (1) \textbf{instruction hallucination}—text-only chain-of-thought (CoT) reasoning cannot fully prevent factual errors due to inherent information bottlenecks; (2) \textbf{reward hacking}—dynamic policy optimization against static reward models allows agents to exploit flaws in reward functions. To address these issues, we propose JarvisEvo, a unified image editing agent that emulates an expert human designer by iteratively editing, selecting appropriate tools, evaluating results, and reflecting on its own decisions to refine outcomes. JarvisEvo offers three key advantages: (1) an interleaved multimodal chain-of-thought (iMCoT) reasoning mechanism that enhances instruction following and editing quality; (2) a synergistic editor–evaluator policy optimization (SEPO) framework that enables self-improvement without external rewards, effectively mitigating reward hacking; and (3) support for both global and local fine-grained editing through seamless integration of Adobe Lightroom. 
On ArtEdit-Bench, JarvisEvo outperforms Nano-Banana by an average of \textbf{18.95\%} on preservative editing metrics, including a substantial \textbf{44.96\%} improvement in pixel-level content fidelity. Project page: https://jarvisevo.vercel.app/.

\end{abstract}    
\section{Introduction}
\label{sec:intro}


\begin{figure*}[t]
    \centering
    \setlength{\abovecaptionskip}{0.1cm} 
    \setlength{\belowcaptionskip}{-0.2cm}
    \includegraphics[width=0.96\textwidth]{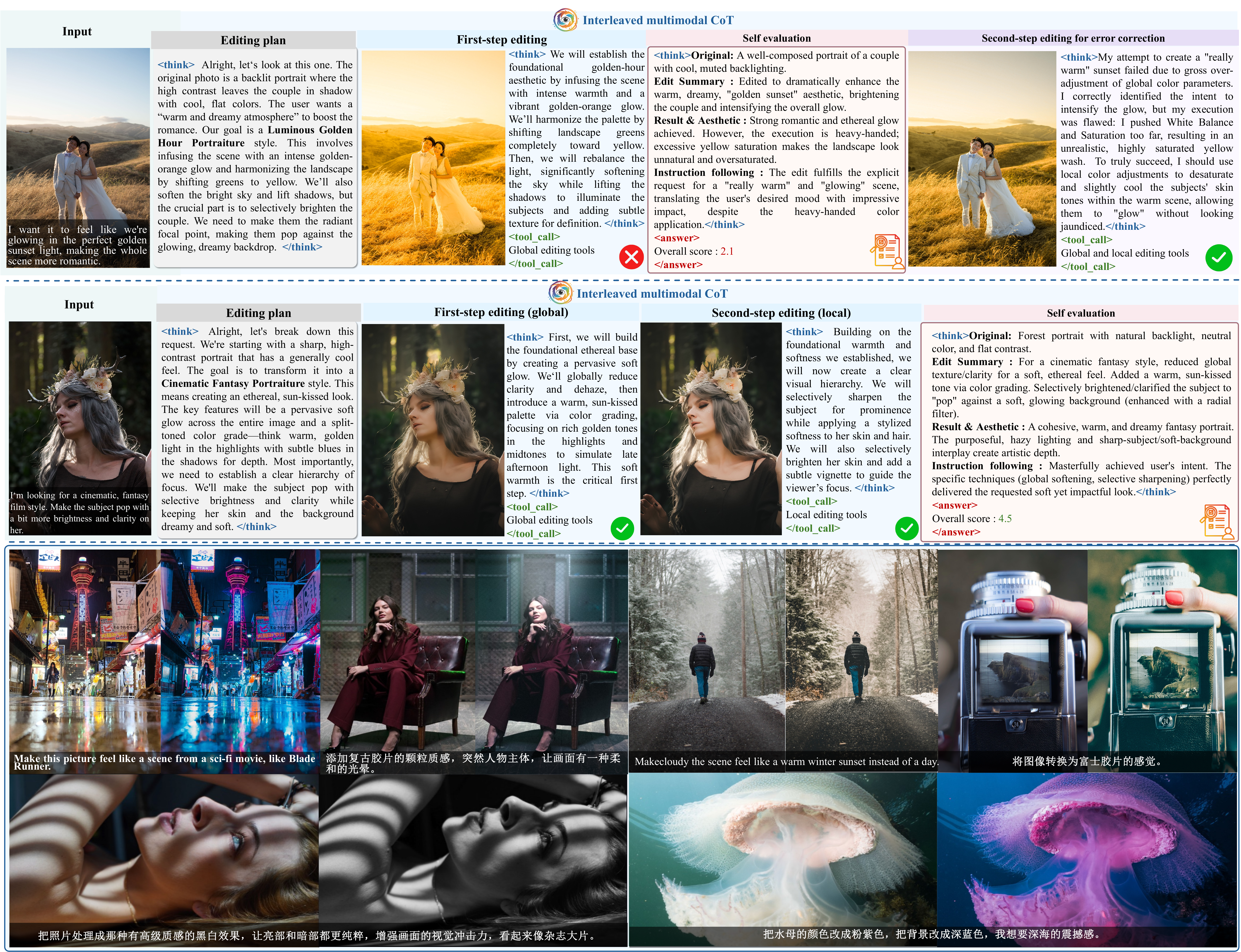}
    \caption{JarvisEvo performs interleaved multimodal Chain-of-Thought (iMCoT) reasoning for image editing, which marries multi-step planning, dynamic tool orchestration, and iterative visual feedback. This closed-loop workflow incorporates self-evaluation and refinement to ensure the final output is both visually compelling and faithful to the creative vision.}
    \label{fig:teaser}
    \vspace{-0.4cm}
\end{figure*}

Recent advances in instruction-based image editing models~\citep{hurst2024gpt, team2023gemini, deng2025bagel, xie2025show, lin2025uniworld, lin2025jarvisart, chen2025photoartagent, dutt2025monetgpt} have enabled the creation of images with remarkable fidelity and diversity, thereby expanding the boundaries of human creative expression. The latest cutting-edge generative models, such as GPT-Image-1~\citep{hurst2024gpt}, Nano-Banana~\citep{team2023gemini}, and Qwen-Image-Edit~\citep{wu2025qwen}, demonstrate outstanding capability in creative composition synthesis. Another research line develops agentic tool-integrated editors, such as JarvisArt~\citep{lin2025jarvisart}, PhotoArtAgent~\citep{chen2025photoartagent}, and MonetGPT~\citep{dutt2025monetgpt}, which leverage preservative tools like Adobe Lightroom for fine-grained, expert-level refinement.

Despite their promise, these models face two fundamental challenges that limit their reliability. \textbf{The first is instruction hallucination}, which occurs when processing complex or lengthy user prompts, causing outputs to diverge from expectations. Recent works~\citep{deng2025bagel, team2024chameleon, fang2025got, duan2025got, jiang2025t2i, wang2025promptenhancer} incorporate textual Chain-of-Thought (CoT) reasoning to enhance instruction comprehension, achieving notable gains in image generation and long-term reasoning tasks. However, this text-centric approach suffers from a critical information bottleneck: the model encodes the source image only once at the beginning and cannot interact with or verify intermediate edited images during reasoning process. As a result, the model's textual hypotheses about image manipulations remain unverified against actual visual feedback, preventing the model from ensuring that ongoing edits align with user intent.
\textbf{The second challenge is reward hacking}, which occurs when aligning models with human preferences through reinforcement learning (RL)~\citep{pan2022effects,clark2023directly,xu2023imagereward,lee2023aligning,ba2025enhancing,flux1kreadev2025}. This issue stems from a fundamental mismatch: the reward model remains static throughout RL, while the policy model is continuously updated. This discrepancy allows the policy to exploit flaws in the reward signal to achieve high scores—without genuinely improving editing capability. Offline reward calibration methods~\citep{xu2023imagereward, wu2023human, kirstain2023pick, zhang2024learning, ma2025hpsv3} attempt to align reward functions with diverse models and aesthetic demands before RL. However, these approaches are resource-intensive, require large-scale human preference annotations, and fail to address the core limitation: the reward model remains fixed and cannot update during RL training.

To address these challenges, we introduce JarvisEvo, a unified photo editing agent that jointly serves as both editor and evaluator to enable self-improvement. As illustrated in Figure~\ref{fig:teaser}, JarvisEvo operates like a skilled designer—editing step by step, evaluating intermediate results, adaptively selecting appropriate tools, and self-correcting errors—to produce coherent, intent-aligned results. JarvisEvo has three key merits: \textbf{(1) Interleaved Multimodal Chain-of-Thought (iMCoT)}: Unlike text-only CoT, which depends on hypothetical outcomes articulated in language, iMCoT leverages visual reasoning to directly observe the consequences of edits, establishing a closed perception–action loop. The model engages in a multimodal dialogue with itself—generating text to hypothesize, creating images to test these hypotheses, and then using further text to reflect on visual outcomes and determine subsequent actions. This dynamic interplay enhances context-aware, accurate decision-making. \textbf{(2) Synergistic Editor–Evaluator Policy Optimization (SEPO)} consists of two co-adaptive optimization loops. (i) The editor loop utilizes self-evaluation scores to generate intrinsic rewards during RL, eliminating reliance on external reward models, while a selective loss masking (SLM) mechanism prevents information leakage from the self-evaluation context. (ii) The evaluator loop continuously refines the model’s evaluation capability using human-annotated assessment data. This loop regularizes the editor’s updates, effectively curbing self-deception and reward hacking. Through this co-evolutionary design, JarvisEvo evolves as both a self-improving editor and a trustworthy evaluator. \textbf{(3) On-policy Reflection Data Generation}: During SEPO, the editor-generated rollout trajectories are automatically converted into reflective training samples. These samples are then used to fine-tune the model, enabling the model to develop self-reflective capabilities.
Our contributions can be summarized as follows:
\begin{itemize}
\item We introduce JarvisEvo, a unified photo editing agent that integrates both editing and evaluation within a single model. JarvisEvo operates like an expert designer—observing visual feedback, reasoning over intermediate outcomes, adaptively selecting tools, and refining results through a closed perception–action loop.
\item We introduce SEPO, a co-evolutionary optimization framework that jointly refines JarvisEvo’s editing and evaluation abilities. SEPO employs a self-reward and SLM mechanism to optimize the editor and adopts verifiable reinforcement learning to refine the evaluator, thereby ensuring stable self-evolution while avoiding self-deceptive or exploitative reward strategies.
\item Experiments on ArtEdit-Bench show that JarvisEvo outperforms the leading agent-based editing model JarvisArt and the commercial editing model Nano-Banana in fine-grained editing tasks. Additionally, its evaluation capabilities surpass those of expert models, aligning more closely with human judgment.
\end{itemize}

\section{Related work}
\label{sec:related_work}
\noindent\textbf {Reinforcement learning.} Reinforcement learning (RL) plays a crucial role in aligning (multimodal) large language models ((M)LLMs)~\citep{Jaech2024Openai,liu2025visualrft,huang2025visionr1,yang2025r1,shen2025vlm} and visual generation systems~\citep{jiang2025t2i,duan2025got,xue2025dancegrpo,liu2025flow} with human preferences. Recent advancements in this domain fall into three paradigms: learning from human feedback (RLHF), with verifiable rewards (RLVR), and from internal feedback (RLIF).
Despite their advantages, each approach exhibits significant limitations. RLHF~\citep{ouyang2022training,rafailov2023direct,touvron2023llama,gao2023scaling} requires costly and static reward models that are vulnerable to reward hacking, policy misalignment, and poor generalization to out-of-distribution data. RLVR~\citep{guo2025deepseek,liu2023your,code-r1,hu2025open,lambert2024tulu,xiaomi2025mimo,team2025kimi} eliminates learned rewards but is restricted to tasks with explicit verification criteria, limiting its applicability to open-ended scenarios such as aesthetic assessment or instruction following. RLIF~\citep{zhao2025learning,zuo2025ttrl,liu2025spark,zhao2025absolute,zelikman2022star,poesia2024learning,yuan2024self}, meanwhile, often suffers from self-deception and overconfidence, which can trigger rapid training collapse by reinforcing incorrect behavior.
To overcome these challenges and promote robust self-improvement, we introduce a hybrid framework—Synergistic Editor-Evaluator Policy Optimization (SEPO)—which combines RLIF and RLVR. SEPO leverages RLIF’s self-supervised rewards with RLVR’s verifiable feedback in a dual-loop framework, enabling stable and grounded self-improvement.

\noindent\textbf{Reasoning paradigm.} A growing body of recent work has incorporated text-centric chain-of-thought (CoT) reasoning into visual generation, improving image quality~\citep{deng2025bagel, team2024chameleon, fang2025got, duan2025got, jiang2025t2i, wang2025promptenhancer}.
Despite these advances, this purely text-based CoT approach introduces an inherent information bottleneck~\citep{su2025thinking,zheng2025deepeyes} in image editing tasks. Specifically, in this paradigm, images are first encoded as static representations, followed by textual CoT to assume the editing steps, and finally the model directly generates the edited image. However, the lack of visual feedback during the reasoning process causes the model to struggle with aligning the assumed visual operations with the actual stepwise edits, leading to deviations between the edited image and the user target result. To address this issue, inspired by OpenAI-o3's ``think-with images" paradigm~\citep{xu2025visual,wang2025multimodal,zhao2025cot,zheng2025deepeyes,zhang2025thyme,chern2025thinking,lai2025mini,fu2025refocus,wang2025pixelthink}, we propose integrating visual and textual reasoning at each editing step, emulating the iterative retouching process of human designers. By introducing visual feedback, the model can validate intermediate results and adjust its decisions according to the image's evolving state, thus improving decision-making accuracy and minimizing the risk of hallucinated or off-target edits.

\noindent\textbf{Instruction-based editing.} It has become a cornerstone of modern artificial intelligence, allowing machines to modify visually appealing and semantically coherent content based on text prompts. Recent unified image editing models have achieved significant advancements in both comprehension and creative content generation. Noteworthy examples include closed-source models such as GPT-4o~\citep{hurst2024gpt} and Nano-Banana~\citep{team2023gemini}, as well as open-source models like Bagel~\citep{deng2025bagel}, Show-o2~\citep{xie2025show}, and Uniworld-V1~\citep{lin2025uniworld}. Additionally, agentic tool-integrated editing models demonstrate impressive comprehension and decision-making capabilities, emulating human designer workflows and coordinating various editing tools for fine-grained aesthetic refinement. Representative works in this area include JarvisArt~\citep{lin2025jarvisart}, PhotoArtAgent~\citep{chen2025photoartagent}, and MonetGPT~\citep{dutt2025monetgpt}. Despite these breakthroughs, two major limitations persist: (1) the system's susceptibility to hallucinations triggered by complex user instructions, despite incorporating textual reasoning, and (2) the reliance on fixed reward models in reinforcement learning, which remains vulnerable to reward hacking.
\section{Method}

\begin{figure}[t]
    \centering
    \setlength{\abovecaptionskip}{0.05cm} 
    \includegraphics[width=1\columnwidth]{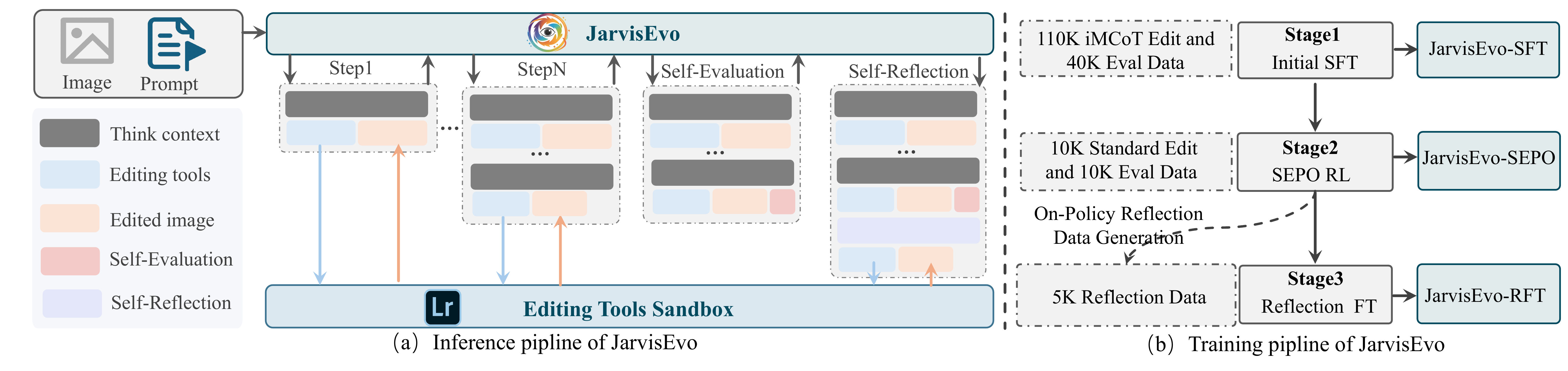}
    \caption{Inference and training pipelines of JarvisEvo.
    }
    \label{fig:piplines}
    \vspace{-0.3cm}
\end{figure}

\subsection{Overview}
\label{sec:Overview}
Figure~\ref{fig:piplines} illustrates the inference and training pipelines of JarvisEvo. 
During inference (Figure~\ref{fig:piplines}(a)), given a source image and user instruction, JarvisEvo engages in an iMCoT reasoning process of multi-step editing, self-evaluation, and self-reflection to produce results that align with both aesthetic standards and user intent. For training (Figure~\ref{fig:piplines}(b)), JarvisEvo follows a three-stage pipeline: Stage 1 performs iMCoT supervised fine-tuning (Sec.~\ref{SFT}) on 150K labeled instances (110K editing, 40K evaluation) to establish multimodal reasoning, tool use, and self-evaluation. Stage 2 employs Synergistic Editor-Evaluator Policy Optimization (Sec.~\ref{sec:SPRO}) on standard data (10K editing, 10K evaluation) to move beyond imitation toward autonomous learning, developing the model into a proficient editor and reliable evaluator. Stage 3 applies Reflection Fine-Tuning (Sec.~\ref{sec:R_data}) on 5K on-policy reflection samples generated during SEPO, strengthening error detection and self-correction.

\subsection{Training Recipe}\label{sec:framework}
\subsubsection{Cold-Start Supervised Fine-tuning}\label{SFT}
We employ supervised fine-tuning to initialize JarvisEvo as both an editor and an evaluator, with four key objectives: (1) to instill the grammar of multimodal reasoning; (2) to enable preliminary generation of interleaved visual and textual content; (3) to develop competence in selecting editing tools based on intermediate visual cues; and (4) to facilitate self-evaluation of aesthetic quality and instruction adherence.

\subsubsection{Synergistic Editor-Evaluator Policy Optimization}\label{sec:SPRO}
\begin{figure*}[t]
    \centering
    \includegraphics[width=1\linewidth]{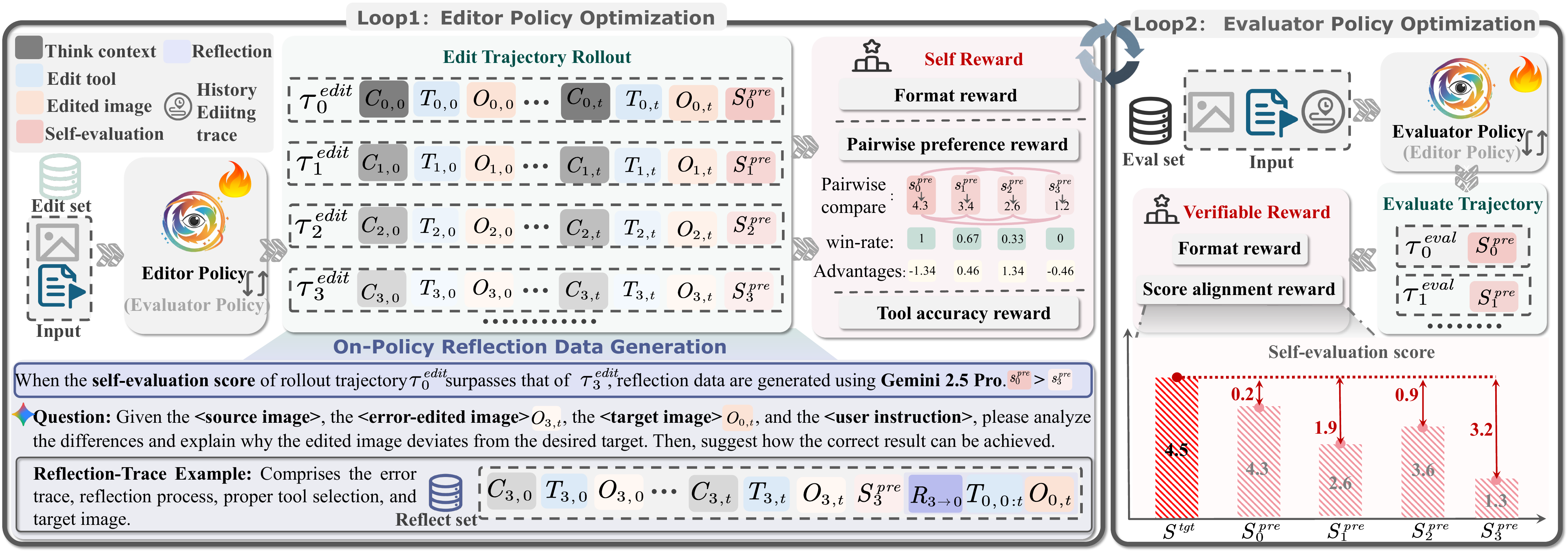}
    \caption{The Synergistic Editor–Evaluator Optimization (SEPO) framework consists of two iterative loops. \textbf{Loop 1} optimizes the editor policy using self-evaluation scores, thereby improving iMCoT reasoning and tool use. In addition, an \textit{\textbf{online reflection data generation pipeline}} autonomously constructs reflection samples, which are then used to further fine-tune the model’s reflective capabilities. \textbf{Loop 2} refines the evaluator policy with human-labeled evaluation data to ensure reliable assessment and to mitigate self-deception or reward hacking during editor optimization.}
    \label{fig:framework}
    \vspace{-0.3cm}
\end{figure*}

To advance JarvisEvo beyond imitation toward autonomous learning, we propose the Synergistic Editor-Evaluator Policy Optimization (SEPO) RL framework. As depicted in Figure~\ref{fig:framework}, SEPO integrates two iterative optimization loops:
\textbf{(1) Editor-policy optimization}, where the model self-rewards using internal evaluation scores without external reward models to refine its editing strategy; and
\textbf{(2) Evaluator-policy optimization}, where the same model is trained on a human-annotated evaluation dataset containing verifiable scores of aesthetic quality and instruction adherence. This loop strengthens the model’s evaluative ability while mitigating risks such as self-deception and reward hacking that may arise during editor optimization. Through this dual-loop design, JarvisEvo evolves concurrently as a self-improving editor and a reliable evaluator.

\noindent\ding{202}\noindent\textbf{Editor policy optimization loop}. 
As illustrated in Figure~\ref{fig:framework}, given an input image $I$ and a user query $Q$, the model first generates textual reasoning content $C$, interleaved with tool invocations $T$, which are executed in a sandboxed external environment. The execution yields an edited image $O$, which is concatenated into the ongoing reasoning trajectory. This loop continues until the model produces a final self-evaluation content $S$, assessing whether the output image meets the desired aesthetic quality and aligns with the user’s intent. The overall process can thus be represented as:
\begin{equation}
    \tau^{edit} = \{(I, Q); ([C_0, T_0, O_0], \dots, [C_t, S^{pred}])\},
    \label{eq:editing_trajectory}
\end{equation}
where $t$ denotes the maximum number of interaction rounds for the sample.

\noindent\textbf{Reward design.} The overall reward for editor policy optimization, denoted as \(R_{\text{edit}}\in[0, 3]\) comprises three components: \(R_{\text{edit}} = R_f + R_{ta} + R_{pp}\): 
\begin{itemize}
    \item \textbf{Format reward} (\(R_f\in[0, 1]\)): enforces correctly structured outputs using the designated tags (\texttt{<think>}, \texttt{<tool\_call>}, \texttt{<answer>}).
    \item \textbf{Tool accuracy reward} (\(R_{ta} \in [0, 1]\)): measures the correctness of the tool name and parameter settings, ensuring valid operations and numerical values. Details are in the supplementary.
    \item \textbf{Pairwise preference reward} (\(R_{pp}\in[0, 1]\)): quantifies the aesthetic quality and instruction compliance of the edited image. For each input \((I, Q)\), the editor policy samples \(G\) trajectories \(\{\tau^{edit}_0, \dots, \tau^{edit}_{G-1}\}\), each with a self-evaluation score \(s^{pred}\) from $S^{pred}$. We then convert these absolute scores into a relative reward by computing the trajectory’s win rate:
    \(R_{pp}(\tau^{edit}_i) = \frac{1}{G-1} \sum_{j \neq i} \mathbb{I}(s^{pred}_i > s^{pred}_j)\), where \(\mathbb{I}(\cdot)\) denotes the indicator function. By using pairwise comparisons, the reward reflects preference relations among trajectories and yields more stable optimization than raw scalar scores.
\end{itemize}
\noindent\textbf{Optimization.} We train the editor policy using the Group Relative Policy Optimization (GRPO) objective~\citep{guo2025deepseek}:
\begin{equation}
\begin{gathered}
\mathcal{J}_{\mathrm{GRPO}}(\theta)=\mathbb{E}_{(I, Q) \sim D,\left\{\tau_i^{\text {edit }}\right\}_{i=1}^G \sim \pi_\theta(\cdot \mid I, Q)}
{\left[\frac{1}{\sum_{i=1}^G\left|\tau_i^{\text {edit }}\right|} \sum_{i=1}^G \sum_{j=1}^{\left|\tau_i^{\text {edit }}\right|}\left(A_{i, j}\right)\right],}
\end{gathered}
\label{eq:grpo}
\end{equation}
where \(\left| \tau^{edit} i \right|=\sum{k=0}^{t_i}{\left| C_{i,k} \right|}+\sum_{k=0}^{t_i-1}{\left| T_{i,k} \right|}\) denotes the total number of tokens generated by the editor for artistic reasoning and tool invocation. \(A_{i,j}\) is the advantage computed from the final trajectory rewards \(\{r_1, r_2, \ldots, r_G\}\) within the same group \(A_{i,j} = \frac{r_i - \text{mean}(\{r_i\}_{i=1}^G)} {\text{std}(\{r_i\}_{i=1}^G)} \quad \text{for each token } j \text{ in } \tau^{edit}_i.\) The reward \(r_i\) evaluates the overall editing trajectory quality, considering iMCoT reasoning structure, tool-use accuracy, aesthetic appeal, and instruction compliance. To maintain stable optimization, inspired by AdaCot~\citep{lou2025adacot}, we introduce selective loss masking (SLM), excluding self-evaluation tokens \(S^{pred}\) from the loss computation to prevent information leakage and avoid training collapse during editor policy learning. Unlike the original GRPO formulations, we omit the KL penalty term against a reference model. This design choice encourages the model to more freely adapt its behavior to our custom response format and structured reward signals. In practice, we observe that this leads to faster convergence and comparable performance, while also simplifying the training pipeline.

\noindent\ding{203}
\textbf{Evaluator policy optimization loop}.
In this loop, we optimize the evaluator policy using a human-annotated evaluation set. Each training example is a quadruple \(\langle I, Q, H, S^{tgt} \rangle\), where $I$ is the source image, $Q$ is the user query, \(H = \{[C_0, T_0, O_0], \dots, [C_t]\}\) denotes the editing trajectory, 
and \(S^{tgt}\) is the human-annotated evaluation content (judgments and scores). The evaluator is given \(\langle I, Q, H\rangle\) and produces a self-evaluation \(S^{pre}\). We define the associated evaluation trajectory $\tau^{eval}$ as:
\begin{equation}
\tau^{eval} = \{(I, Q, H); (S^{pred})\}.
\label{eq:evaluation_trajectory}
\end{equation}
This context-preserving design ensures that the evaluator receives inputs aligned with the editor's generation context, thereby maintaining input domain consistency across the two optimization loops.


\noindent\textbf{Reward design.} The overall reward for evaluator policy optimization, denoted as \(R_{\text{eval}}\in[0, 2]\), comprises two components: \(R_{\text{eval}} = R_f + R_{sa}\):
\begin{itemize}
    \item \textbf{Format reward} (\(R_f\in[0, 1]\)) ensures structural adherence by enforcing placement of rationales within \texttt{<think>} tags and scores within \texttt{<answer>} tags, yielding standardized, parsable outputs.
    \item \textbf{Score alignment reward} (\(R_{sa}\in[0, 1]\)) measures consistency between predicted and human-annotated scores. Given input \((I, Q, H)\), the evaluator samples \(G\) trajectories \(\{\tau^{eval}_1, \dots, \tau^{eval}_G\}\), each producing a score \(s^{pre}_i\) in \(S^{pre}\). The reward is computed as:
    \(
    R_{sa}(\tau^{eval}_i) = \exp \left( -\frac{1}{2} \left( \frac{|s^{pre}_i - s^{tgt}_i|}{\sigma} \right)^2 \right) + \epsilon,
    \)
    where \(\sigma=0.5\) controlling error tolerance and \(\epsilon\) is a small positive constant to ensure non-zero gradients. This encourages score alignment with human judgments.
\end{itemize}

\noindent\textbf{Optimization.}  
We adopt the same optimization objective defined in Equation~\ref{eq:grpo}, with two key modifications:  
(i) trajectories are generated by sampling self-evaluations \(S^{pre}\) conditioned on the editing context (see Equation~\ref{eq:evaluation_trajectory});  
(ii) the effective token count \(|\tau_i| = \sum_{k=0}^{t_i-1} |S^{pre}_{i,k}|\) includes only tokens from self-evaluation outputs.  
This design constrains the loss to evaluative content, thereby encouraging the model to produce more reliable, unbiased, and human-aligned self-assessments.

\begin{figure}[t]
    \centering
    \setlength{\abovecaptionskip}{-0.05cm} 
    \setlength{\belowcaptionskip}{-0.4cm}
    \includegraphics[width=\linewidth]{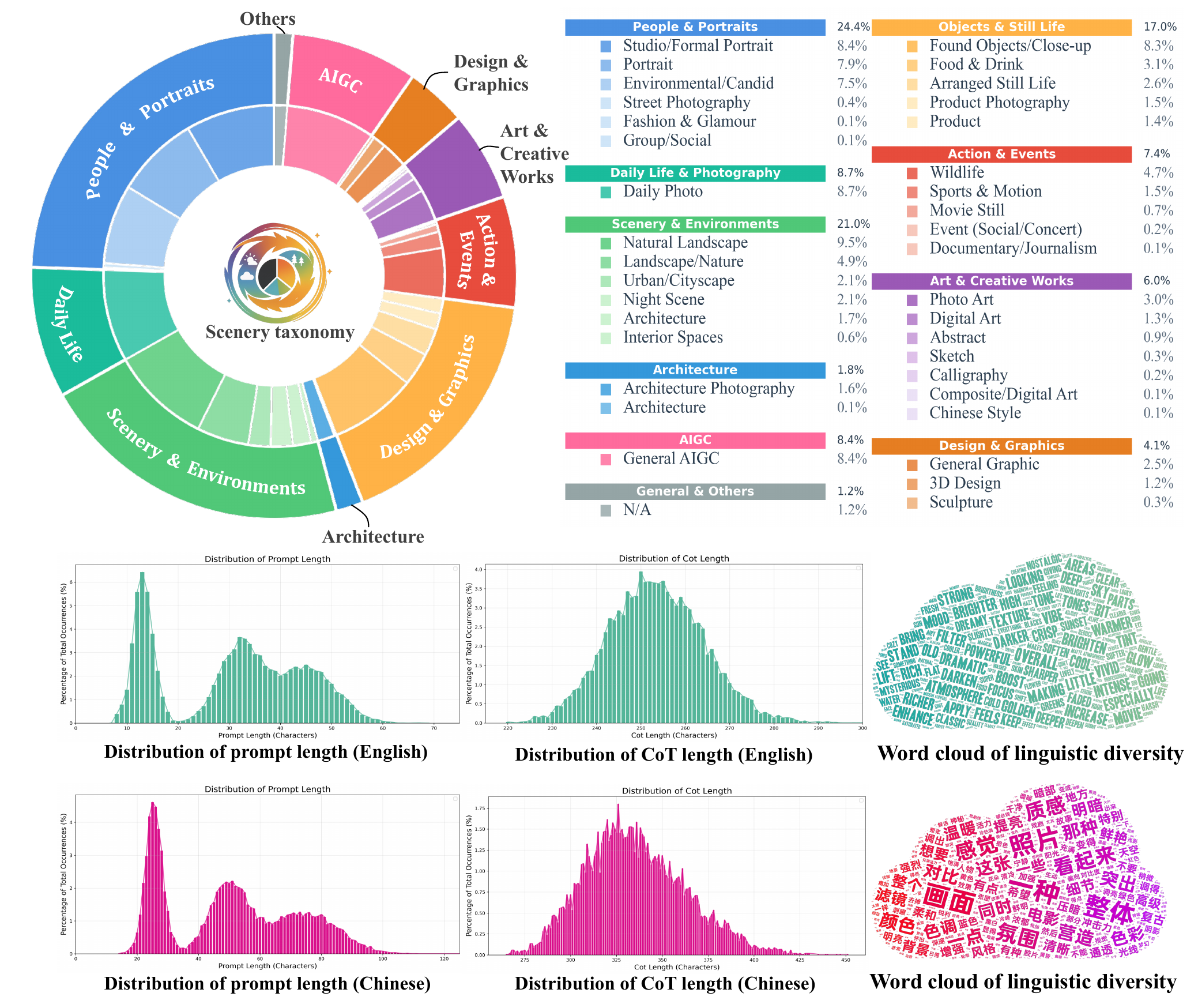} 
    \caption{Scenario and prompt distribution on ArtEdit.}    \label{fig:data_distribution}
    \vspace{-0.2cm}
\end{figure}

\begin{figure*}[!t]
    \centering
    \setlength{\abovecaptionskip}{0.02cm} 
    \setlength{\belowcaptionskip}{-0.3cm}
    \includegraphics[width=\linewidth]{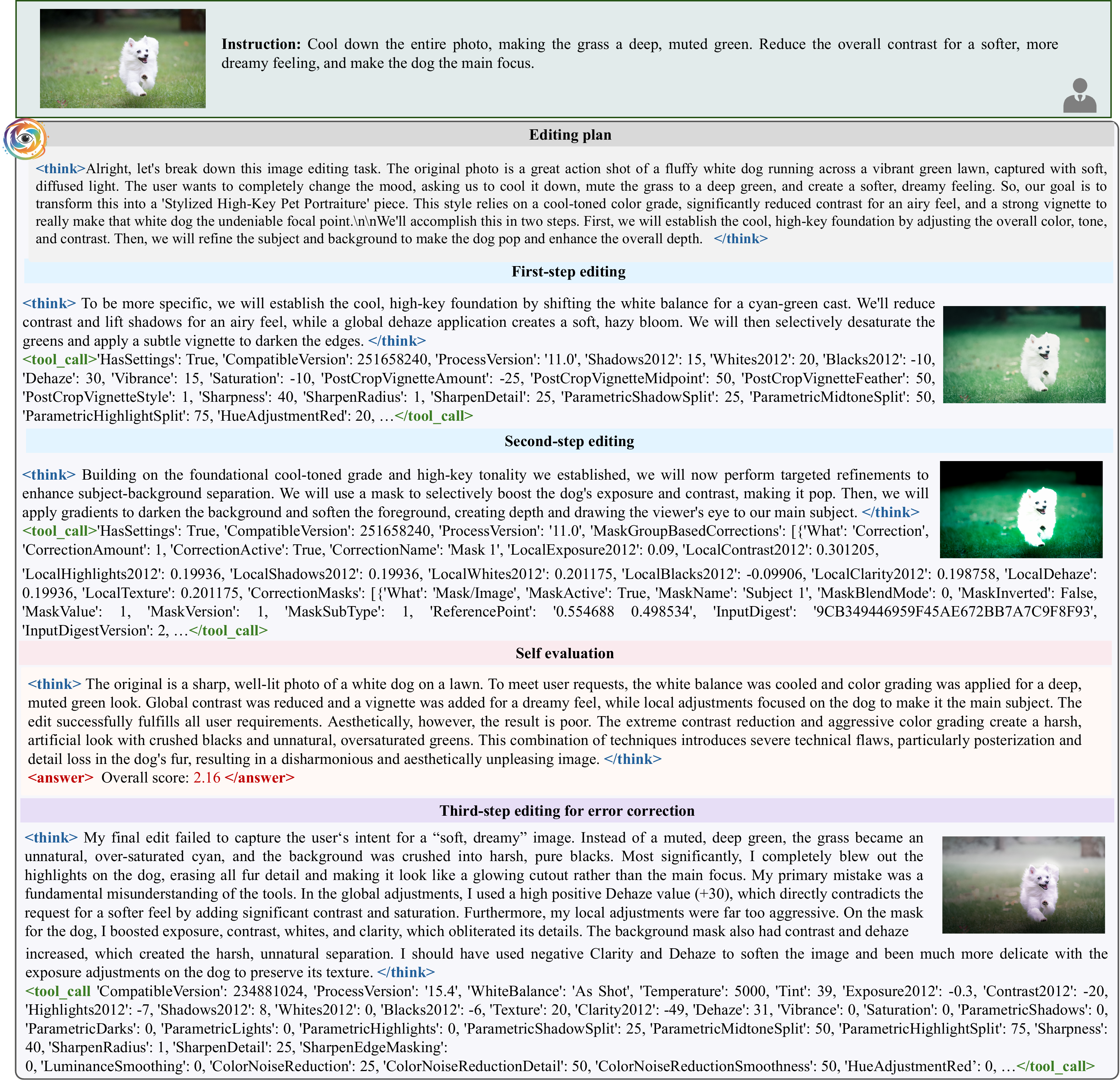}
    \caption{Reflection data sample generated during on-policy updates in SEPO, including erroneous editing trajectories, corrective reflections, accurate editing operations, and corresponding images.}
    \label{fig:reflect_result2}
\end{figure*}

\subsubsection{Reflection Data Generation and Fine-tuning}\label{sec:R_data}
\begin{figure*}[htbp]
    \centering
    \includegraphics[width=1\linewidth]{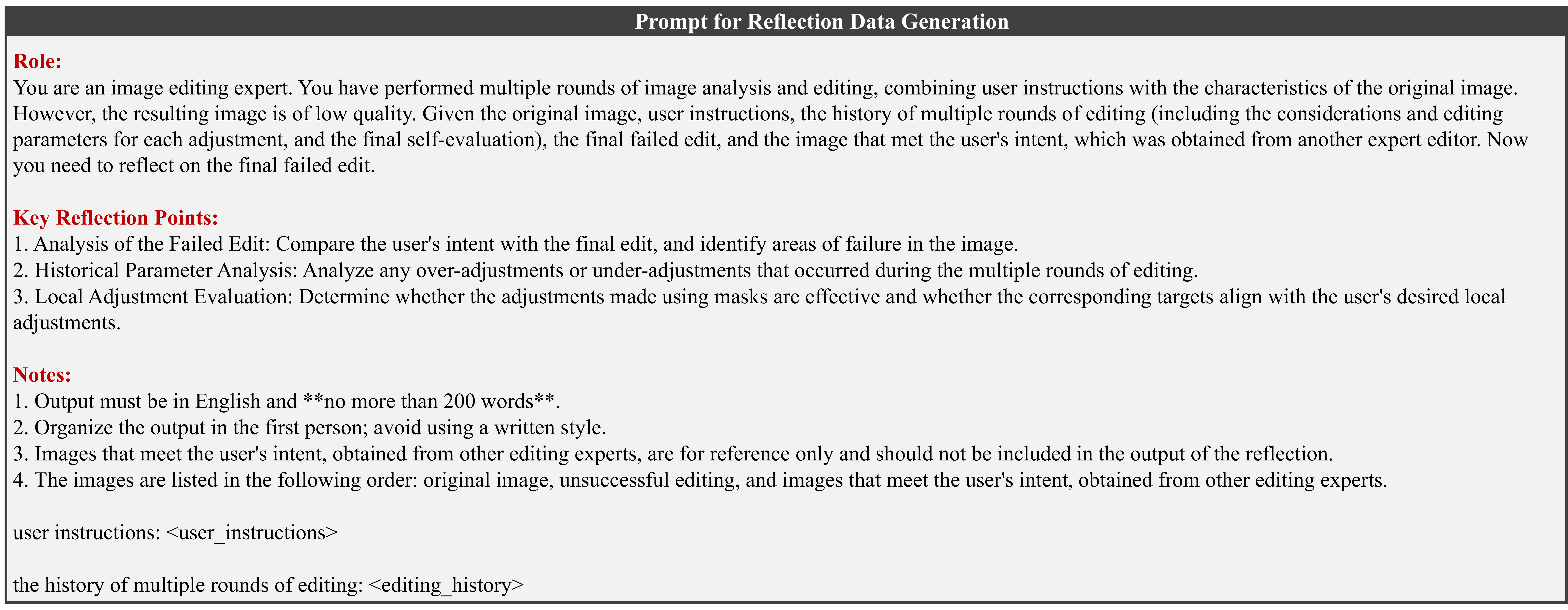}
    \caption{Role-playing prompt utilized to generate on-policy reflection data during the SEPO stage.}
    \label{fig:p_reflection}
\end{figure*}

To enable JarvisEvo to develop self-reflective abilities, we introduce an autonomous, on-policy pipeline that generates reflection trajectories for fine-tuning. As shown in Figure~\ref{fig:framework}, this reflection pipeline is activated during the editor policy optimization loop whenever a trajectory’s self-evaluation score \(s^{(pred)}_0\) surpasses that of another, \(s^{(pred)}_3 \) (i.e., \( s^{(pred)}_0 > s^{(pred)}_3 \)). In such cases, Gemini-2.5-Pro~\citep{team2023gemini} is prompted with \(\langle I, Q, O_3, O_0 \rangle\) to produce a reflection rationale \( R_{3 \to 0} \). The role-play prompt used for this process is shown in Figure~\ref{fig:p_reflection}. The tool configuration \( T_{0,0:t} \) from trajectory \( \tau_0 \) is regarded as the correct editing path, and the corresponding output image \( O_{0,t} \) serves as the target. These elements together define a reflection trajectory: \(\tau^{reft} = \{(I, Q); ([H_3, S^{pre}_3], [R_{3 \to 0}, T_{0,0:t}, O_{0,t}])\},\) where \( H_3 = [C_{3,0}, T_{3,0}, O_{3,0}], \dots, [C_{3,t}] \) captures the erroneous editing history of \( \tau_3 \). The collection of such reflection trajectories forms a training dataset \(D_{reft}\) for supervised fine-tuning. Figure~\ref{fig:reflect_result2} shows a constructed reflection data sample, which includes the error editing trajectory and the corrected edit path.

\begin{figure*}[t!]
    \centering
    \setlength{\abovecaptionskip}{0.03cm} 
    \setlength{\belowcaptionskip}{-0.3cm}
    \includegraphics[width=1\linewidth]{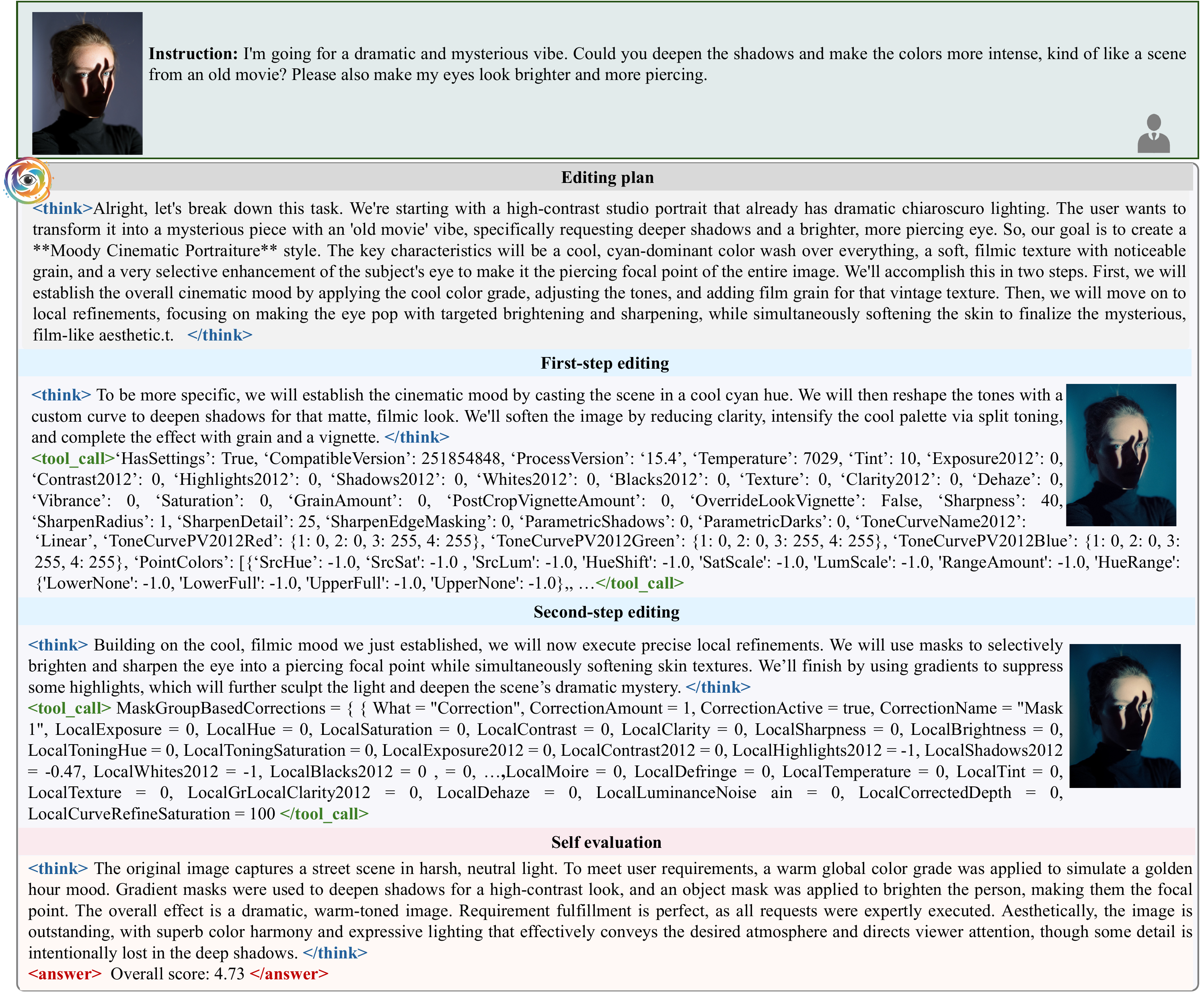}
    \caption{Representative data sample from ArtEdit-Lr, illustrating a fine-grained editing task.}
    \label{fig:data_sample_lrc}
\end{figure*}

\subsection{Training Data}\label{sec:mmart}
We introduce ArtEdit, a comprehensive bilingual (English and Chinese) dataset comprising 170K samples that includes global and local editing tasks, along with their corresponding evaluation data. The dataset is organized into two subsets: ArtEdit-Lr (120K samples) for preservative editing and ArtEdit-Eval (50K samples) for assessment tasks. The source data are collected from public sources~\citep{liang2021ppr10k, lin2025jarvisart, fivek, schuhmann2022laion, qian2025picobanana400klargescaledatasettextguided} and curated by professional photographers.

\noindent\textbf{Editing dataset.}
Figure~\ref{fig:data_distribution} illustrates the distribution of editing prompts and scenarios. ArtEdit comprises 10 scene categories (over 37 subcategories) such as portraits, landscapes, architecture, wildlife, still life, urban scenes, and design. Each data entry consists of (1) a source–target image pair, (2) a user instruction, and (3) an iMCoT editing trace that captures the application of over 200 Lightroom tools, along with the corresponding intermediate edited images.

\noindent\textbf{Evaluation dataset.}
ArtEdit-Eval is uniformly sampled from ArtEdit-Lr to ensure balanced coverage across diverse editing scenarios. Each sample is annotated with an overall floating-point score (1–5) that reflects both aesthetic quality and instruction adherence. Each entry includes (1) a source–target image pair, (2) a user instruction, (3) the complete iMCoT editing trajectory, and (4) human-verified annotation scores. 

\subsubsection{Data Generation Pipeline}
In the editing data subset, each sample is represented as a triplet \(\langle I_{src}, I_{tgt}, Q, H \rangle\) where $I_{src}$ is the source image, $I_{tgt}$ is the target image, $Q$ is the user query, and \(H = \{[C_0, T_0, O_0], \dots, [C_t]\}\) denotes the multimodal editing trajectory. Here, $C$ represents the textual reasoning content, $T$ specifies the editing tool orchestration, and $O$ corresponds to intermediate edited images.
In the evaluation data subset, each sample follows a similar format: \(\langle I_{src}, I_{tgt}, Q, H, S \rangle\), with the addition of $S$, a score evaluating the overall result based on both aesthetic quality and instruction alignment. The data generation pipeline comprises four primary stages: (1) collection and generation of initial image pairs, (2) synthesis of user instructions, (3) multimodal CoT annotation, (4) evaluation data annotation, and (5) quality assurance via data filtering. This pipeline is inspired by recent advancements in agent-based editing models~\citep{lin2025jarvisart}.

\noindent\textbf{Stage 1: Initial iamge pairs with tool configuration collection}. We source high-quality source images from publicly available datasets, such as PPR10K~\citep{liang2021ppr10k}, MMArt~\citep{lin2025jarvisart}, CADB~\citep{zhang2021image}, RPCD~\citep{nieto2022understanding}, FiveK~\citep{fivek}, AesMMIT~\citep{AesExpert}, LAION-Aesthetics~\citep{schuhmann2022laion} and 
pico-banana-400K~\citep{qian2025picobanana400klargescaledatasettextguided}.
While some datasets provide source–target image pairs along with the associated Lightroom tool configurations, others contain only source images. For the latter, we develop a systematic pipeline to synthesize the corresponding target images. Specifically, we first use Gemini-2.5-Pro~\citep{team2023gemini} to generate candidate user instructions from each source image, creating distinct prompts for preservative and generative editing tasks. Then, we employ JarvisArt~\citep{lin2025jarvisart}, conditioned on these generated instructions, to produce edited images together with their Lightroom parameters. 
Finally, the collected image pairs and their corresponding tool configurations \(\langle I_{src}, I_{tgt}, T\rangle\) undergo a quality assurance procedure consisting of an initial automated screening with Gemini-2.5-Pro~\citep{team2023gemini}, followed by manual spot-checks to ensure data reliability.

\begin{figure*}[p]
    \centering
    \vfill
    \includegraphics[width=1\linewidth]{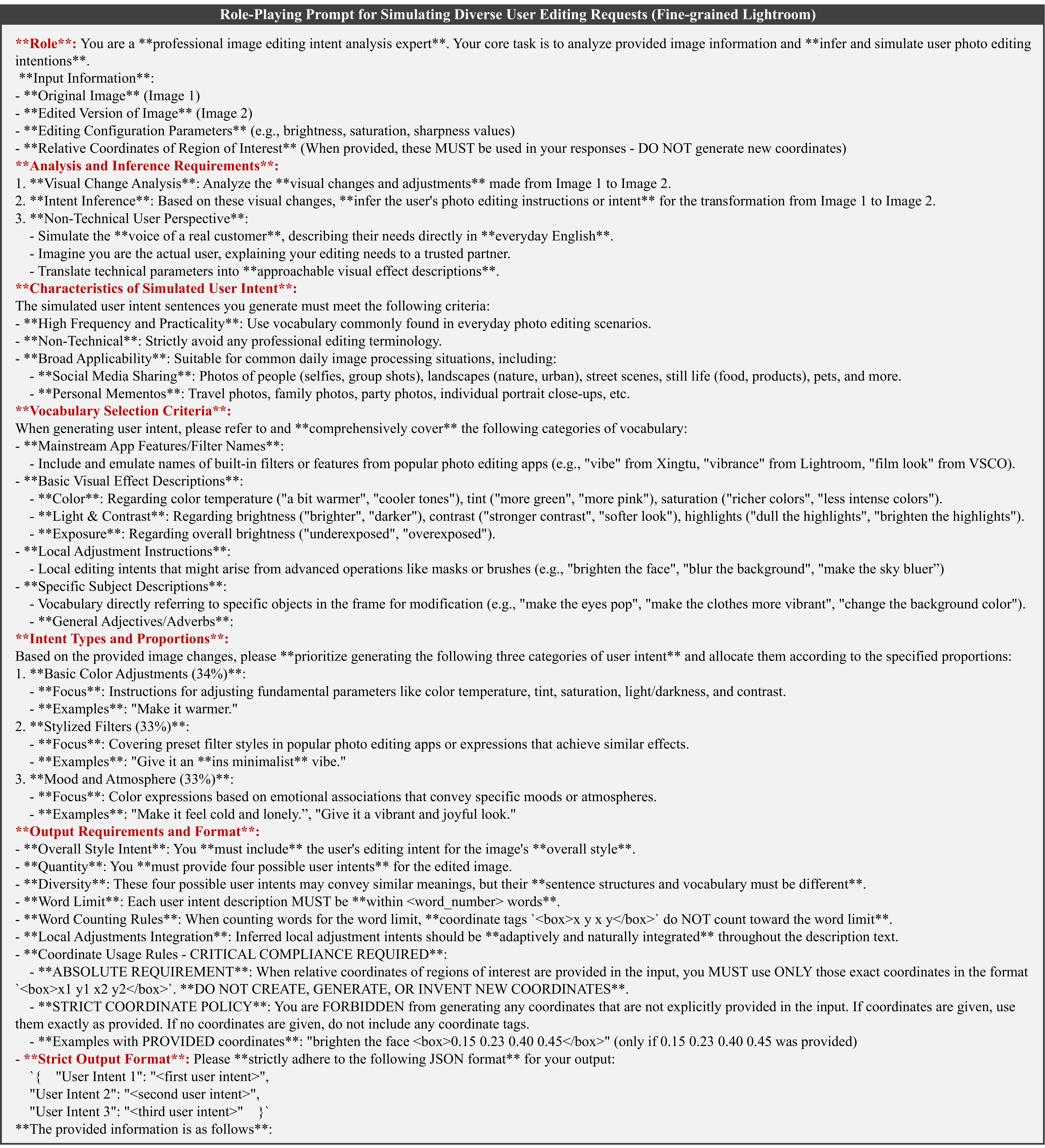}
    \caption{Prompt used to simulate diverse user instructions for fine-grained adjustment tasks.}
    \label{fig:p_instruct}
    \vfill
\end{figure*}

\noindent\textbf{Stage 2: User instructions generation}.
To simulate diverse user editing instructions for different task, we employ Gemini-2.5-Pro~\citep{team2023gemini} with a role-playing prompt to translate each \(\langle I_{src}, I_{tgt} \rangle\) tuple into both fine-grained adjustment and creative editing instructions \(Q\). We generate instructions of varying lengths and complexities to better reflect real-world scenarios and practical requirements. The system prompts used for the fine-grained adjustment task is provided in Prompt~\ref{fig:p_instruct}.

\noindent\textbf{Stage 3: iMCoT annotations generation}. 
The iMCoT annotation process is designed to generate a detailed, step-by-step reasoning trace. Given a sample \(\langle I_{\text{src}}, I_{\text{tgt}}, Q, T\rangle\), where the overalll tool orchestration \(T\) consists of multiple tool-configuration sets \(\{T_1, \dots, T_N\}\), that are used to produce intermediate edited images, we adopt an iterative annotation pipeline.
Rather than generating the entire chain-of-thought at once, we generate a reasoning context \(C_t\) for each editing step \(t\). Gemini-2.5-Pro~\citep{team2023gemini}, acting as an editing analyst expert, is tasked with analyzing the transition from the intermediate image \(I_{t-1}\) to \(I_t\) caused by applying the \(t\)-th step editing tools \(T_t\). To ensure precision, Gemini-2.5-Pro~\citep{team2023gemini} is provided with a comprehensive context at each step, including:
\begin{itemize}
    \item \textbf{Global context:} The user's original instruction \(Q\) and the final target image \(I_{\text{tgt}}\) to maintain alignment with the overall goal.
    \item \textbf{Local dynamics:} The intermediate images before and after the current step (\(I_{t-1}\) and \(I_t\)), along with the specific tool parameters \(T_t\).
    \item \textbf{Historical reasoning context:} The aggregated reasoning from all previous steps \(\{C_0, \dots, C_{t-1}\}\).
\end{itemize}
This strictly guided, step-wise process ensures that the final reasoning trace \(\mathcal{C} = \{C_0, \dots, C_t\}\) is both logically coherent and grounded in the actual visual transformations occurring at each stage of the editing workflow. 

\begin{figure*}[htbp]
    \centering
    \includegraphics[width=1\linewidth]{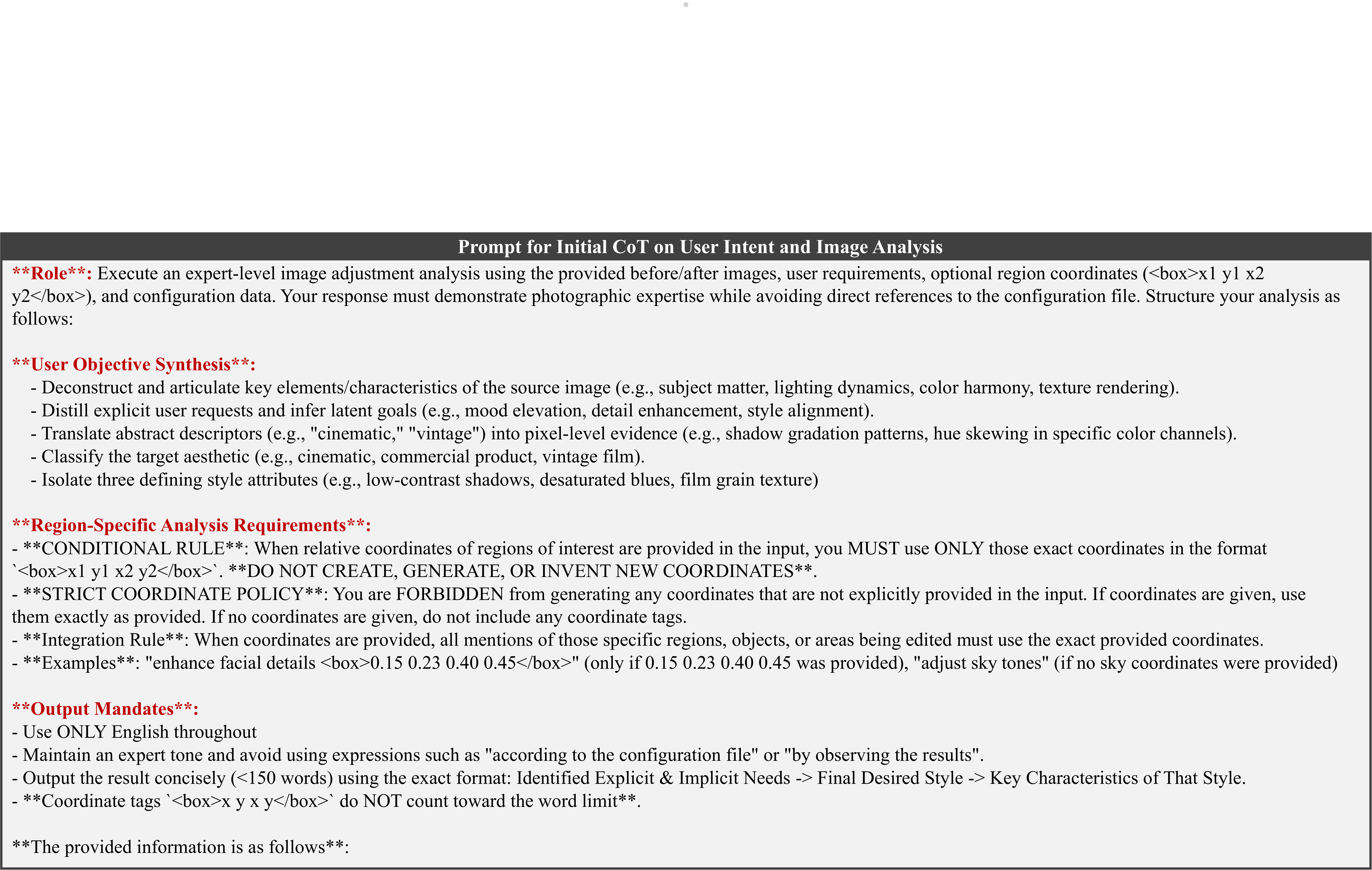}
    \caption{Prompt for generating initial Chain-of-Thought (CoT) annotations, focusing on source image analysis and user intent understanding.}
    \label{fig:p_initial_cot}
\end{figure*}

\begin{figure*}[htbp]
    \centering
    \includegraphics[width=1\linewidth]{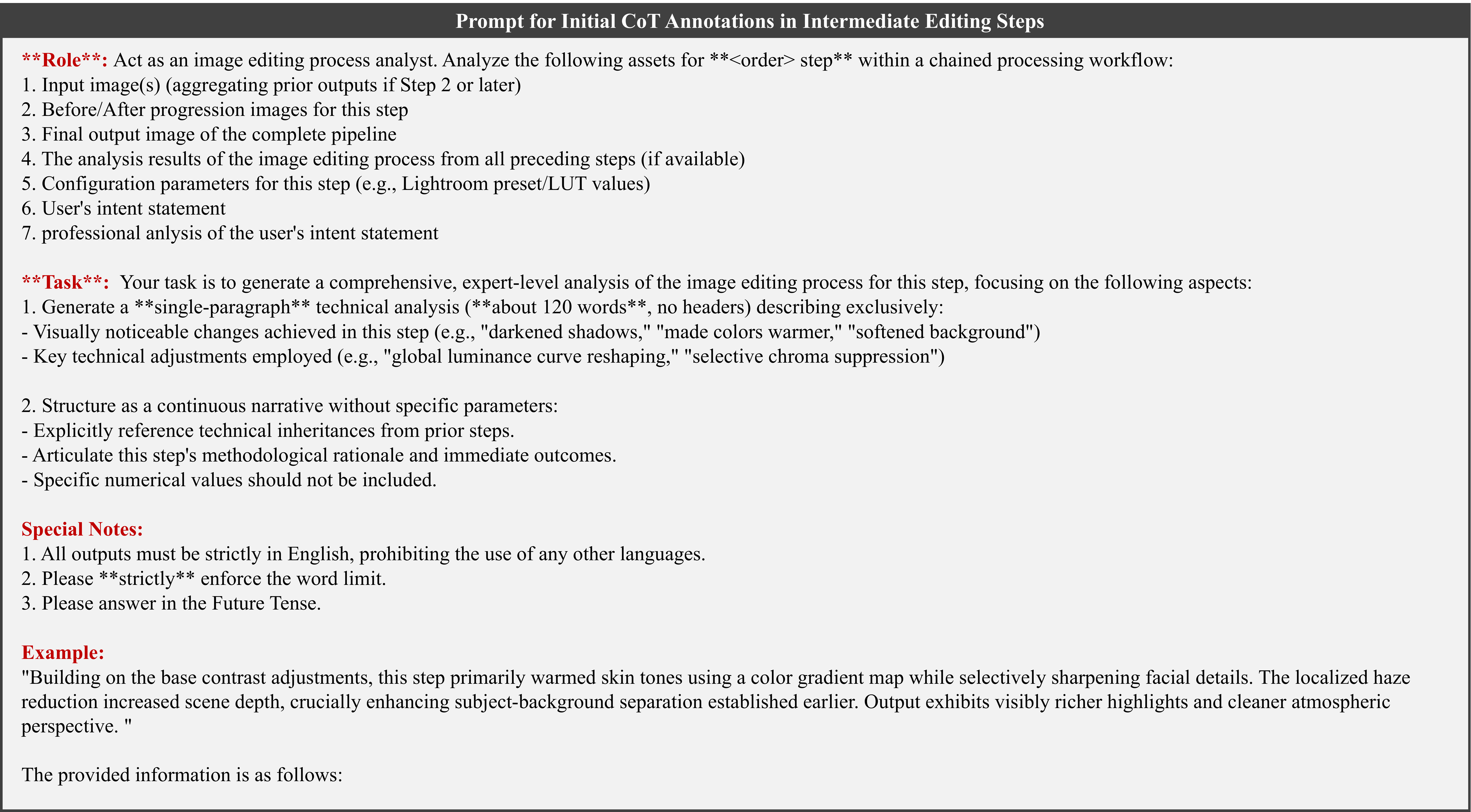}
    \caption{Prompt for generating initial Chain-of-Thought (CoT) annotations for intermediate editing steps.}
    \label{fig:p_initial_cot_steps}
\end{figure*}

\begin{figure*}[htbp]
    \centering
    \includegraphics[width=1\linewidth]{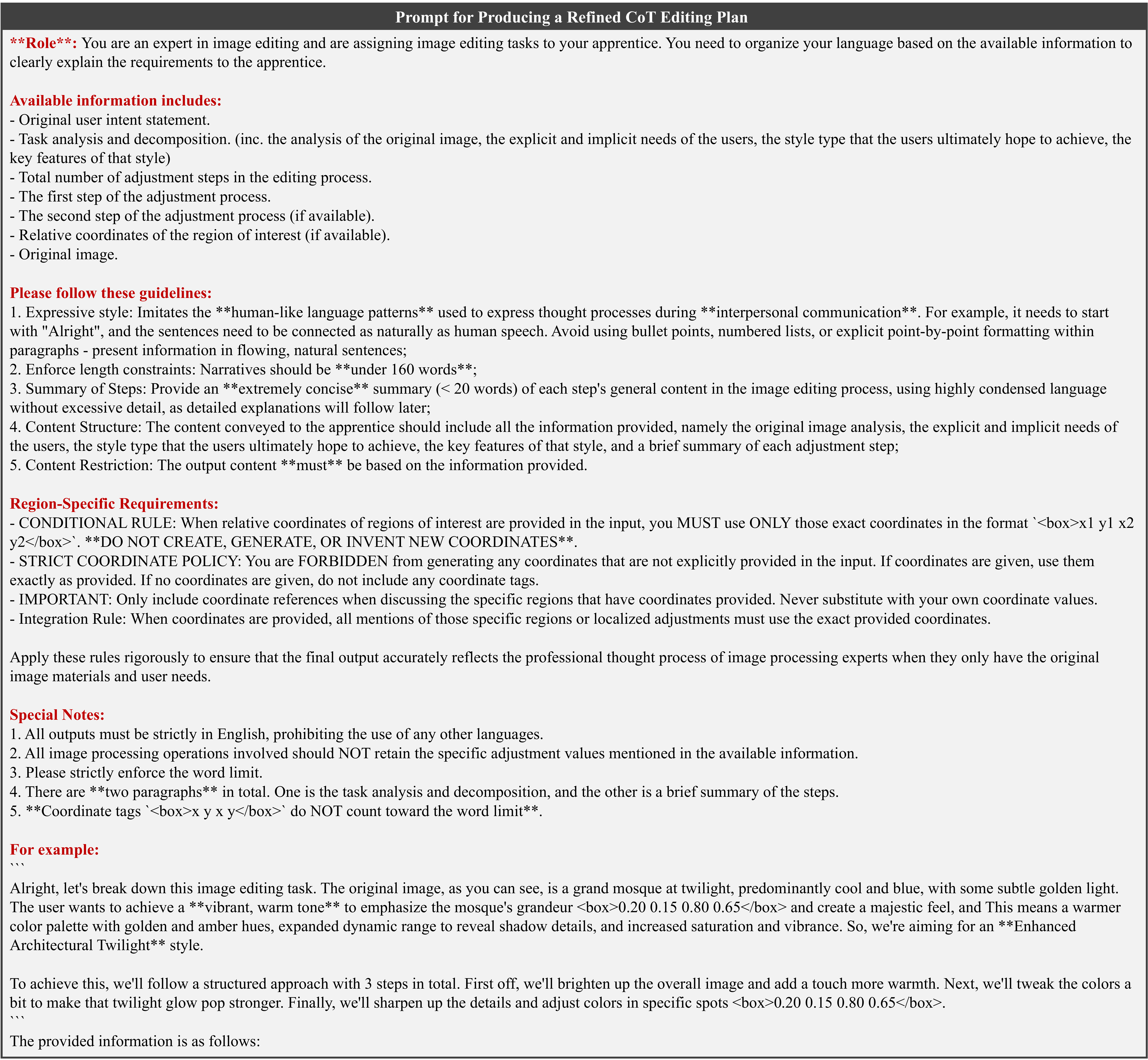}
    \caption{Prompt for producing a refined Chain-of-Thought (CoT) editing plan, incorporating user intent understanding and overall editing strategy.}
    \label{fig:p_refined_cot_lrc}
\end{figure*}

\begin{figure*}[htbp]
    \centering
    \includegraphics[width=1\linewidth]{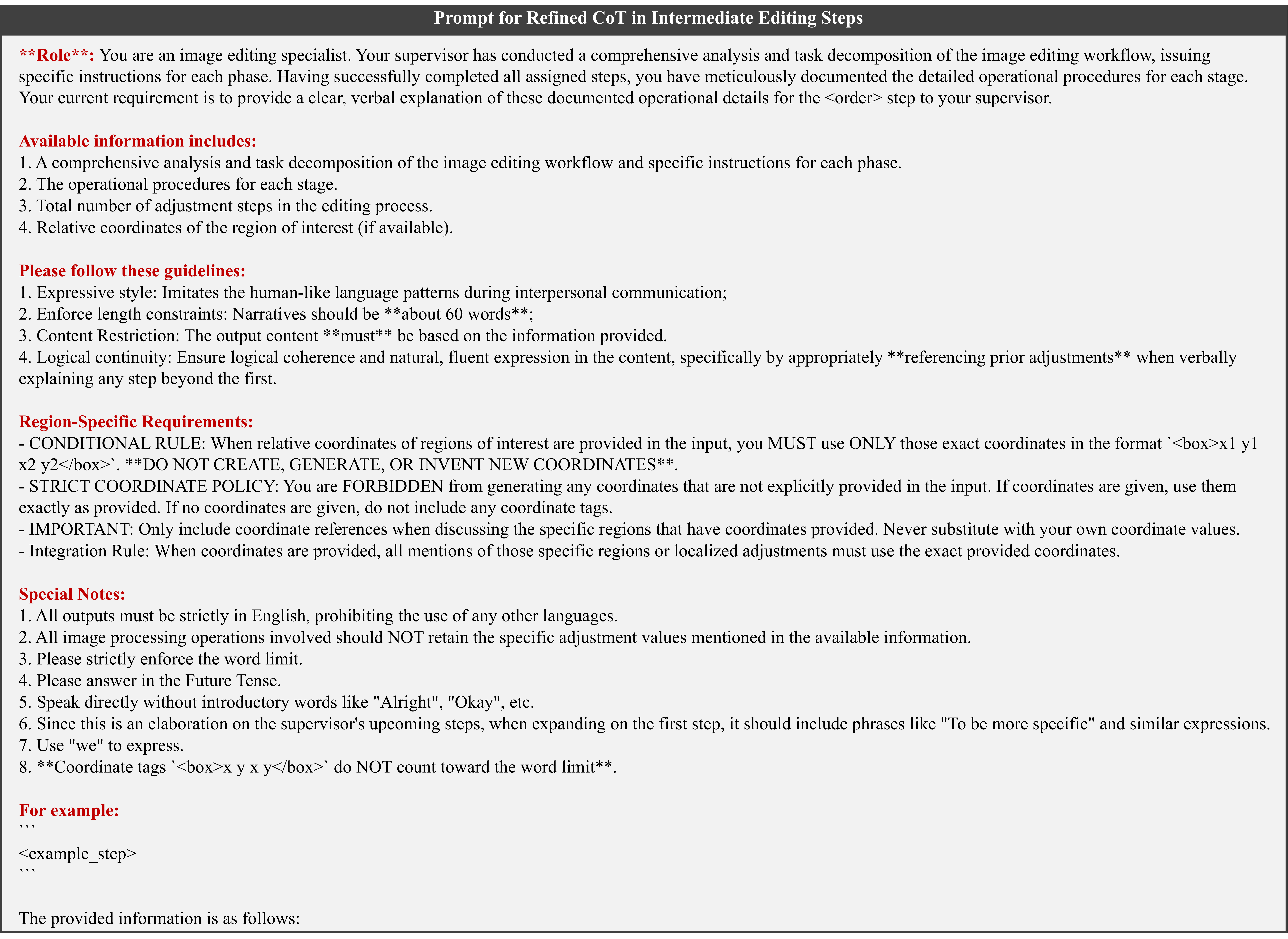}
    \caption{Prompt for generating refined Chain-of-Thought (CoT) annotations for intermediate editing steps.}
    \label{fig:p_refined_cot_step_lrc}
\end{figure*}

We utilize the prompt templates shown in Figures~\ref{fig:p_initial_cot}, \ref{fig:p_initial_cot_steps}, \ref{fig:p_refined_cot_lrc}, and \ref{fig:p_refined_cot_step_lrc} to generate both initial and refined iMCoT annotations. Given that this task involves over 200 parameterized Lightroom tools and often requires multiple editing steps, the resulting transformations are both detailed and complex. To manage this complexity, we implement a two-stage annotation workflow: an initial CoT for capturing first-pass reasoning, followed by a refined CoT that improves accuracy, resolves ambiguities, and ensures consistency across the steps.

\begin{figure*}[htbp]
    \centering
    \includegraphics[width=1\linewidth]{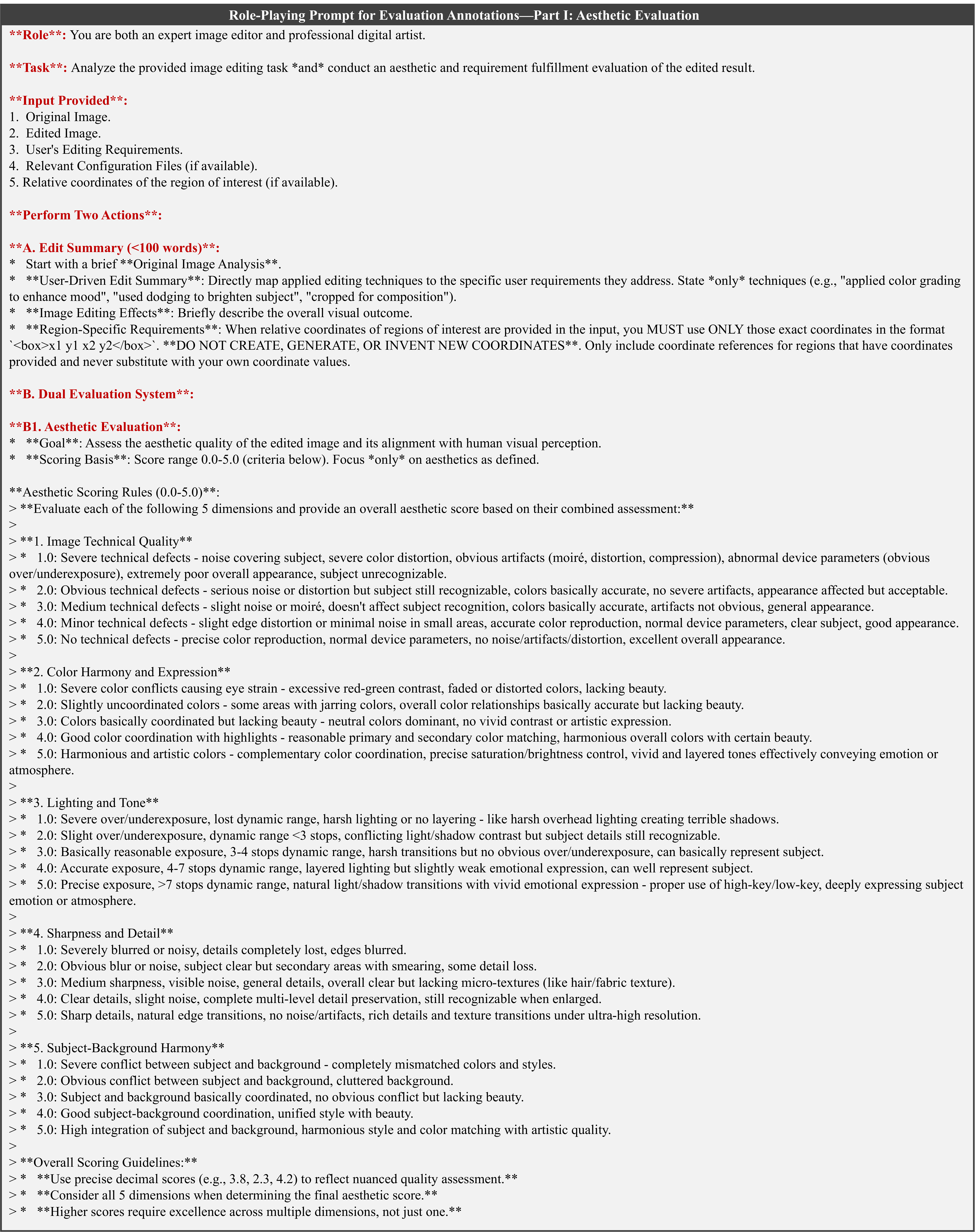}
    \caption{Role-playing prompt for evaluation annotations—Part I: Aesthetic evaluation.}
    \label{fig:p_eval_p1}
\end{figure*}
\begin{figure*}[htbp]
    \centering
    \includegraphics[width=1\linewidth]{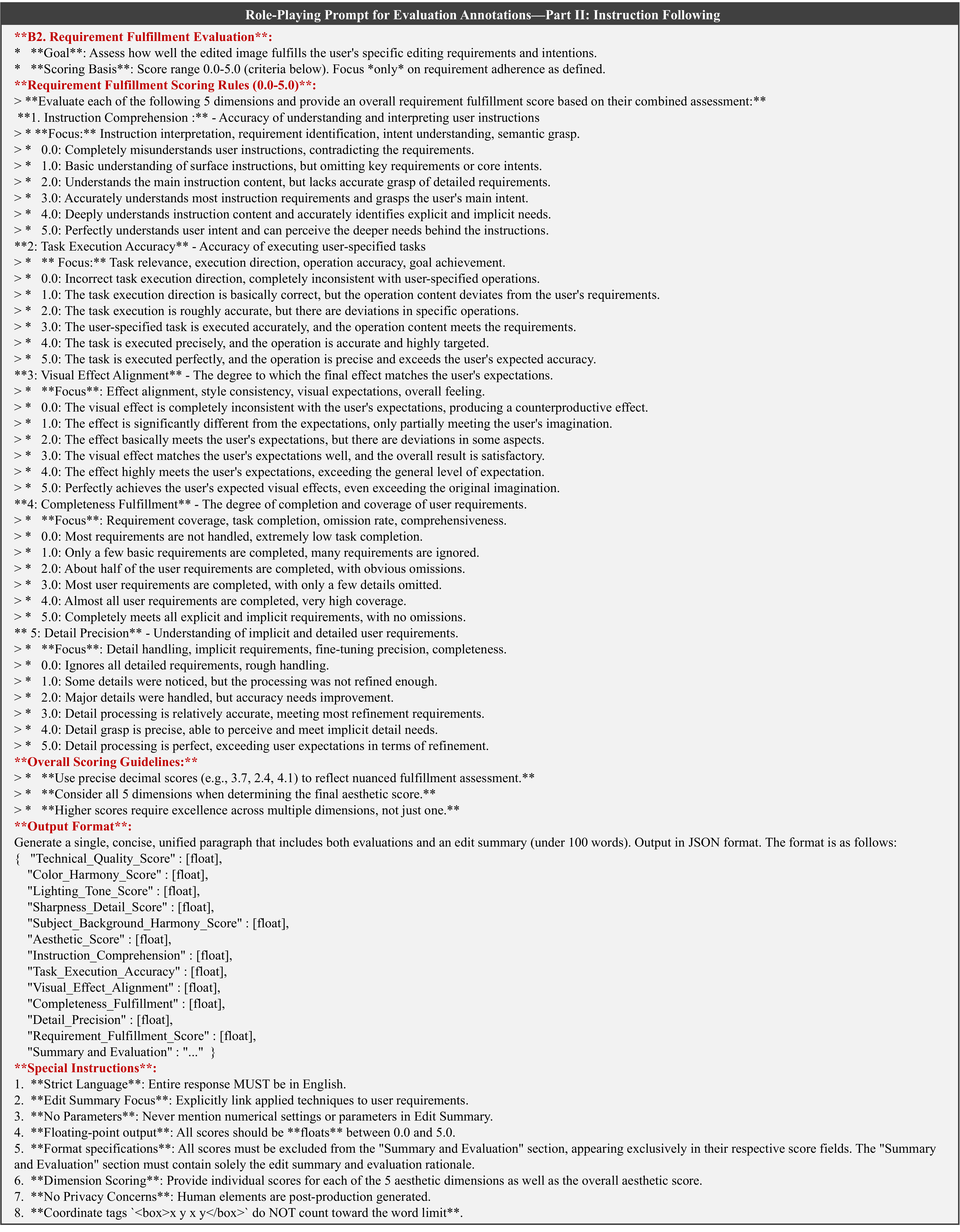}
    \caption{Role-playing prompt for evaluation annotations—Part II: Instruction adherence evaluation.}
    \label{fig:p_eval_p2}
\end{figure*}

\noindent\textbf{Stage 4: Evaluation data annotations}. 
To build the evaluation subset, we use Gemini-2.5-Pro \citep{team2023gemini} to assess each editing sample \(\langle I_{src}, I_{tgt}, Q, H\rangle\). For each sample, Gemini-2.5-Pro generates an evaluation output \(S\), which includes an overall score and a textual rationale based on two criteria: (1) \textit{aesthetic quality}, which evaluates the visual appeal of the target image \(I_{tgt}\) in terms of composition, color harmony, and overall presentation; (2) \textit{Instruction adherence}, which quantifies the fidelity of the editing results against the user intents within $Q$. We implement a role-playing prompting scheme that positions Gemini-2.5-Pro~\citep{team2023gemini} as a professional evaluator following detailed scoring rules. Aesthetic quality is rated along five dimensions—technical quality, color harmony and expressiveness, lighting and tonal coherence, sharpness and detail, and subject–background integration—while instruction adherence is evaluated through comprehension, execution accuracy, visual alignment with instructions, completeness, and fine-detail precision. Each dimension receives a continuous score from 0.0 to 5.0, and the model also provides an overall score and qualitative rationale. The evaluation prompts are shown in Figures \ref{fig:p_eval_p1} and \ref{fig:p_eval_p2}.

\noindent\textbf{Stage 5: Data filtering}. 
To ensure the quality and reliability of the data, our filtering process consists of two stages: automated filtering followed by manual cross-validation. In the first stage, we utilize Gemini-2.5-Pro~\citep{team2023gemini} to automatically assess each sample based on three key criteria: instruction adherence, the aesthetic quality of the edited image, and the consistency of the annotation scores. Gemini-2.5-Pro~\citep{team2023gemini} performs an initial review to detect any discrepancies or potential issues in these areas. In the second stage, manual cross-validation is conducted, where three human reviewers carefully inspect each filtered sample. This step ensures that the automated filtering process aligns with human judgment, validating the accuracy of the annotations and the quality of the final image. This two-tiered approach ensures a robust data filtering process, resulting in high-quality data for model training and evaluation. The filtered editing data is shown in Figure~\ref{fig:data_sample_lrc}.

\begin{table*}[t!]
    \centering
    \setlength{\abovecaptionskip}{-0.01cm} 
    \setlength{\belowcaptionskip}{-0.2cm}
    \caption{Quantitative evaluation of editing performance on ArtEdit-Bench-Lr. The \colorbox{lightpink!90}{best} and \colorbox{lightblue!40}{second-best} results are highlighted. SC, PQ, and O denote the metrics evaluated by GPT-4o~\citep{hurst2024gpt}. Note: Models denoted by $^*$ underperform on the Chinese set of ArtEdit-Bench-Lr due to their limited Chinese language capability.}\label{tab:edit1}
    \scalebox{0.94}{
    \setlength\tabcolsep{4.5pt}
    \renewcommand\arraystretch{0.8}
    \begin{tabular}{lcccccc} 
    \toprule
    Method & Open-Source & $L1_{\times 10^{2}} \downarrow$ & $L2_{\times 10^{3}} \downarrow$ & $\text{SC} \uparrow$ & $\text{PQ} \uparrow$  & $\text{O} \uparrow$ \\
    \midrule
    \multicolumn{7}{c}{\textcolor{gray!100}{English Set}} \\ 
        JarvisArt~\citep{lin2025jarvisart} & \cmark & 13.21 & 38.45 & 7.81 & 8.26 & 7.89 \\
        MonetGPT~\citep{dutt2025monetgpt} & \cmark & \colorbox{lightblue!40}{9.46} & \colorbox{lightblue!40}{17.21} & 6.73 & 8.98 & 7.62 \\
        \midrule
        Instruct-Pix2Pix~\citep{brooks2023instructpix2pix} & \cmark & 14.22 & 41.18 & 6.93 & 7.37 & 7.00 \\
        MagicBrush~\citep{zhang2023magicbrush} & \cmark & 17.11 & 59.68 & 4.96 & 4.23 & 4.29 \\
        OmniGen~\citep{xiao2024omnigen} & \cmark & 25.53 & 109.55 & 4.57 & 2.60 & 3.20 \\
        Step1X-Edit~\citep{liu2025step1x} & \cmark & 11.78 & 29.96 & 4.36 & 5.80 & 4.83 \\
        Bagel~\citep{deng2025bagel} & \cmark & 16.32 & 51.94 & 8.07 & 7.57 & 7.77 \\
        UniWorld-v1~\citep{lin2025uniworld} & \cmark & 11.79 & 29.40 & 8.34 & 8.69 & 8.49 \\
        FLUX-1-Kontext~\citep{batifol2025flux} & \cmark & 19.79 & 78.42 & 8.27 & 8.27 & 8.23 \\
        \midrule
        GPT-Image-1~\citep{hurst2024gpt} & \xmark & 21.20 & 82.77 & \colorbox{lightblue!40}{8.45} & 8.02 & 8.21 \\
        Nano-Banana~\citep{team2023gemini} & \xmark & 11.54 & 28.34 & 8.33 & \colorbox{lightblue!40}{8.92} & \colorbox{lightblue!40}{8.59} \\
        \midrule
        \textbf{JarvisEvo} & \cmark & \colorbox{lightpink!90}{7.82}  & \colorbox{lightpink!90}{12.45} & \colorbox{lightpink!90}{8.53} & \colorbox{lightpink!90}{9.03} & \colorbox{lightpink!90}{8.77} \\
        \midrule
        \multicolumn{7}{c}{\textcolor{gray!100}{Chinese Set}} \\ 
        JarvisArt~\citep{lin2025jarvisart} & \cmark & $-$  & $-$ & $-$ & $-$ & $-$ \\
        MonetGPT~\citep{dutt2025monetgpt} & \cmark & \colorbox{lightblue!40}{9.63} & \colorbox{lightblue!40}{17.49} & 7.01 & 8.99 & 7.81 \\
        \midrule
        Instruct-Pix2Pix$^*$~\citep{brooks2023instructpix2pix} & \cmark & 12.80 & 31.86 & 5.62 & 7.78 & 6.43 \\
        MagicBrush$^*$~\citep{zhang2023magicbrush} & \cmark & 17.11 & 55.30 & 2.78 & 3.00 & 2.57 \\
        OmniGen$^*$~\citep{xiao2024omnigen} & \cmark & 28.55 & 126.39 & 3.05 & 2.38 & 2.41 \\
        Step1X-Edit~\citep{liu2025step1x} & \cmark & 11.64 & 29.27 & 4.47 & 5.68 & 4.86 \\
        Bagel~\citep{deng2025bagel} & \cmark & 17.43 & 58.14 & 7.89 & 7.57 & 7.68 \\
        UniWorld-v1$^*$~\citep{lin2025uniworld} & \cmark & 11.35 & 25.95 & 7.17 & 8.88 & 7.92 \\
        FLUX-1-Kontext$^*$~\citep{batifol2025flux} & \cmark & 10.45 & 22.67 & 5.01 & 8.99 & 6.38 \\
        \midrule
        GPT-Image-1~\citep{hurst2024gpt} & \xmark & 20.42 & 76.23 & \colorbox{lightblue!40}{8.52} & 8.18 & 8.32 \\
        Nano-Banana~\citep{team2023gemini} & \xmark & 11.46 & 27.39 & 8.35 & \colorbox{lightblue!40}{8.99} & \colorbox{lightblue!40}{8.64} \\
        \midrule
        \textbf{JarvisEvo} & \cmark & \colorbox{lightpink!90}{7.63} & \colorbox{lightpink!90}{11.54} & \colorbox{lightpink!90}{8.54} & \colorbox{lightpink!90}{9.04} & \colorbox{lightpink!90}{8.76} \\
        \bottomrule
    \end{tabular}}
    \vspace{-0.2cm}
\end{table*}

\begin{figure*}[t!]
    \centering
    \includegraphics[width=1\linewidth]{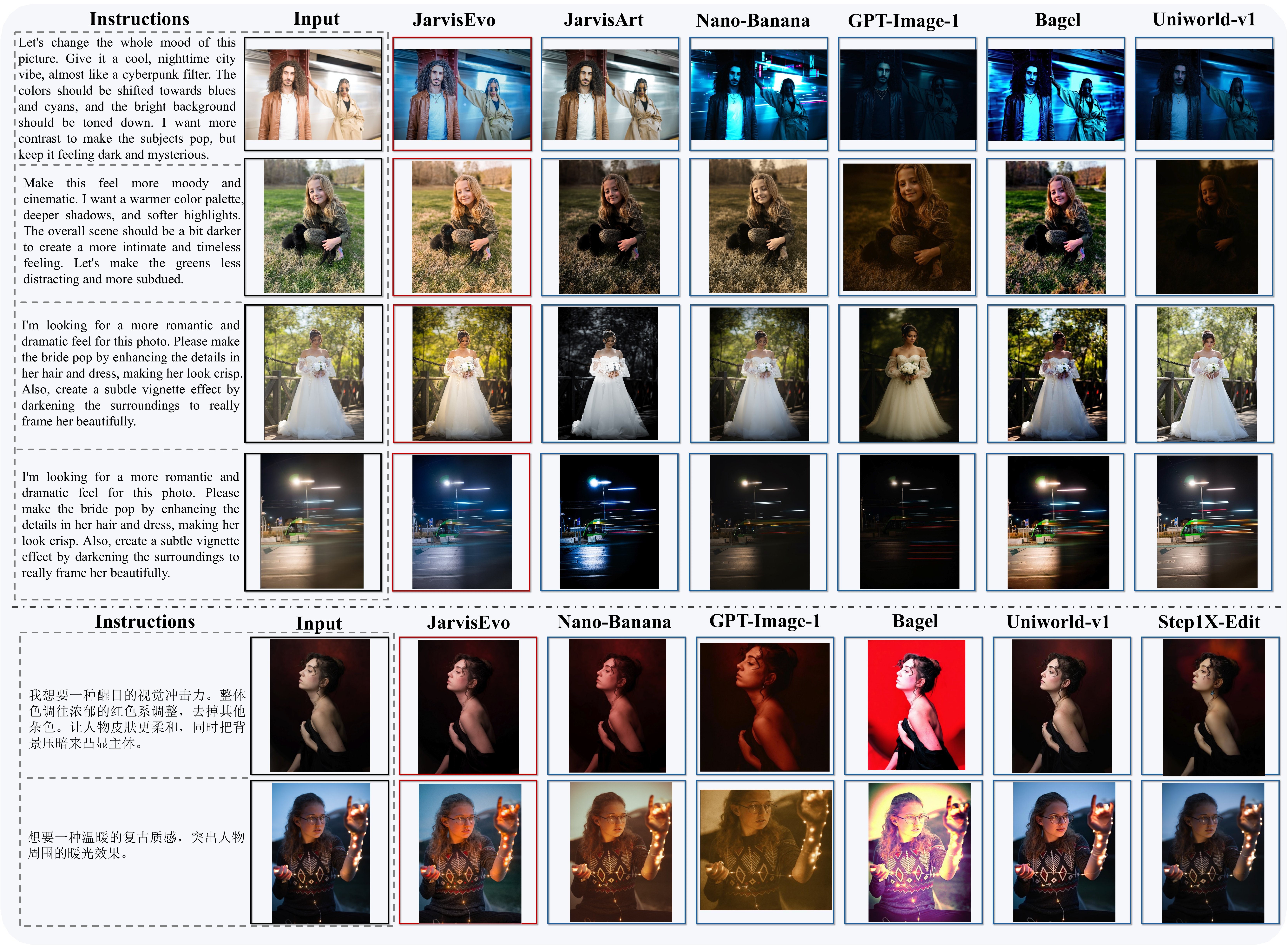}
    \caption{Visual comparison on ArtEdit-Bench. JarvisEvo outperforms other methods in fine-grained retouching due to its Lightroom-based preservative editing.}
    \label{fig:visual_res}
    \vspace{-0.3cm}
\end{figure*}
\section{Experiment}
\subsection{Experimental Setup}
\noindent\textbf{Implementation details.}
JarvisEvo is trained through a three-stage protocol as shown in Figure~\ref{fig:piplines}. \textit{(1) Initial supervised fine-tuning} is based on Qwen3-VL-8B-Instruct~\citep{Qwen2.5-VL}, fine-tuning on 150K annotated examples (110K ArtEdit-LR, 40K ArtEdit-Eval) over 2 epochs, using a batch size of 2 and a learning rate of $1 \times 10^{-5}$, implemented via the Llama-Factory~\citep{zheng2024llamafactory} framework. \textit{(2) SEPO reinforcement learning} is applied via the vlm-r1 framework~\citep{shen2025vlm}, containing two optimization loops: an editing policy trained on 10K standard instruction samples from ArtEdit-LR, and an evaluator policy trained on 10K ArtEdit-Eval samples. This phase is trained for 1 epoch with a batch size of 1, learning rate of $1 \times 10^{-6}$, and produces 4 responses per query. \textit{(3) Reflection Fine-Tuning} is performed on a 5K-sample reflection dataset (collected on-policy in Sec.~\ref{sec:R_data}) for 1 epoch, with a learning rate of $5 \times 10^{-6}$. All experiments are conducted on 32 A100 GPUs. 

\noindent\textbf{Tool settings.} To support professional photo editing, we integrated over 200 Lightroom retouching tools into JarvisEvo via the A2L protocol~\citep{lin2025jarvisart}. The details of the tools are presented in Table~\ref{tab:tools}.

\noindent\textbf{ArtEdit-Bench.} 
We construct ArtEdit-Bench, comprising two subsets: 
(1) ArtEdit-Bench-Lr (800 samples: 400 English, 400 Chinese), selected from the ArtEdit-Lr dataset. It evaluates both global and local fine-grained retouching capabilities. (2) ArtEdit-Bench-Eval (200 English samples), sampled from ArtEdit-Eval, is used to fairly assess the model’s self-evaluation capabilities against assessment models.

\subsection{Experimental Results}
\subsubsection{Performance of Editing}
We compare JarvisEvo with open-source agent-based models (JarvisArt~\citep{lin2025jarvisart}, MonetGPT~\citep{dutt2025monetgpt}), generative methods (InstructPix2Pix~\citep{brooks2023instructpix2pix}, MagicBrush~\citep{zhang2023magicbrush}, OmniGen~\citep{xiao2024omnigen}, UniWorld-v1~\citep{lin2025uniworld}, Step1X-Edit~\citep{liu2025step1x}, FLUX-1-Kontext~\citep{batifol2025flux}, Bagel~\citep{deng2025bagel}), and commercial systems (GPT-Image-1~\citep{hurst2024gpt}, Nano-Banana~\citep{team2023gemini}). We employ six widely used assessment metrics~\citep{wu2025qwen,lin2025jarvisart,zhang2023magicbrush}: L1, L2, SC, PQ, and O. L1 and L2 to measure the average pixel-level absolute difference between the retouched image and reference image. SC evaluates the alignment between the instruction text and the image (0–10 scale). PQ evaluates contextual coherence and artifact presence (0–10 scale).The overall score O is computed as the geometric mean of SC and PQ, averaged across all samples.
As shown in Table~\ref{tab:edit1} and Figure~\ref{fig:visual_res}, JarvisEvo exhibits superior performance in fine-grained retouching tasks. On ArtEdit-Bench-Lr, it outperforms Nano-Banana by an average of 18.95\% across all evaluation metrics for both English and Chinese subsets. In particular, for content fidelity metrics (L1 and L2), JarvisEvo achieves an average improvement of 44.96\% over Nano-Banana~\citep{team2023gemini}.

\begin{table}[!t]
\centering
    \setlength{\abovecaptionskip}{0.1cm} 
\caption{Assessment comparison on ArtEdit-Bench-Eval. Models marked with $^*$ (using fixed prompts) assess only image quality, not instruction following.}
\label{tab:aesthetic_eval}
\setlength\tabcolsep{27pt}
\renewcommand\arraystretch{0.85}
\scalebox{1}{
\begin{tabular}{lcc}
\toprule
Method & SRCC $\uparrow$ & PLCC $\uparrow$ \\
\midrule
\multicolumn{3}{c}{\textcolor{gray!60}{Task-specific IQA MLLM}} \\ 
VisualQuality-R1$^*$~\citep{wu2025visualquality} & 0.5645 & 0.5018 \\
Q-insight$^*$~\citep{li2025q} & 0.5369 & 0.4937 \\
\midrule
\multicolumn{3}{c}{\textcolor{gray!60}{Vallina MLLM}} \\ 
Qwen3-VL-8B~\citep{Qwen2.5-VL} & 0.4659 & 0.4479 \\
Qwen3-VL-256B-A22B~\citep{Qwen2.5-VL} & 0.5706 & 0.5650 \\
Gemini-2.5-Flash~\citep{team2023gemini} & 0.6188 & 0.6441 \\
\midrule
\textbf{JarvisEvo} & 0.7243 & 0.7116 \\
\bottomrule
\end{tabular}
}
\end{table}

\subsubsection{Performance of Evaluation}
We evaluate the self-assessment capability of JarvisEvo on ArtEdit-Bench-Eval. The SRCC~\citep{blondel2020fast} and PLCC~\citep{xu2024boosting} results presented in Table~\ref{tab:aesthetic_eval} reveals that JarvisEvo achieve the best results on average, validating that evaluation capability aligned better with human preferences than task-specific assessment models~\citep{wu2025visualquality} and general MLLMs.
using  and  to measure alignment with human ratings. Compared with aesthetic assessment models (VisualQuality-R1~\citep{wu2025visualquality}, Q-insight~\citep{li2025q}) and general MLLMs (Qwen3-VL-8B~\citep{Qwen2.5-VL}, Qwen3-VL-256B-A22B~\citep{Qwen2.5-VL}, Gemini-2.5-Flash~\citep{team2023gemini}), JarvisEvo achieves the highest and most stable correlations, demonstrating accurate aesthetic and instruction-following evaluation aligned with human preferences.

\begin{wrapfigure}{r}{0.5\textwidth}
    \vspace{-1cm}
    \centering
    \setlength{\abovecaptionskip}{0.02cm} 
    \setlength{\belowcaptionskip}{-0.2cm}
    \includegraphics[width=\linewidth]{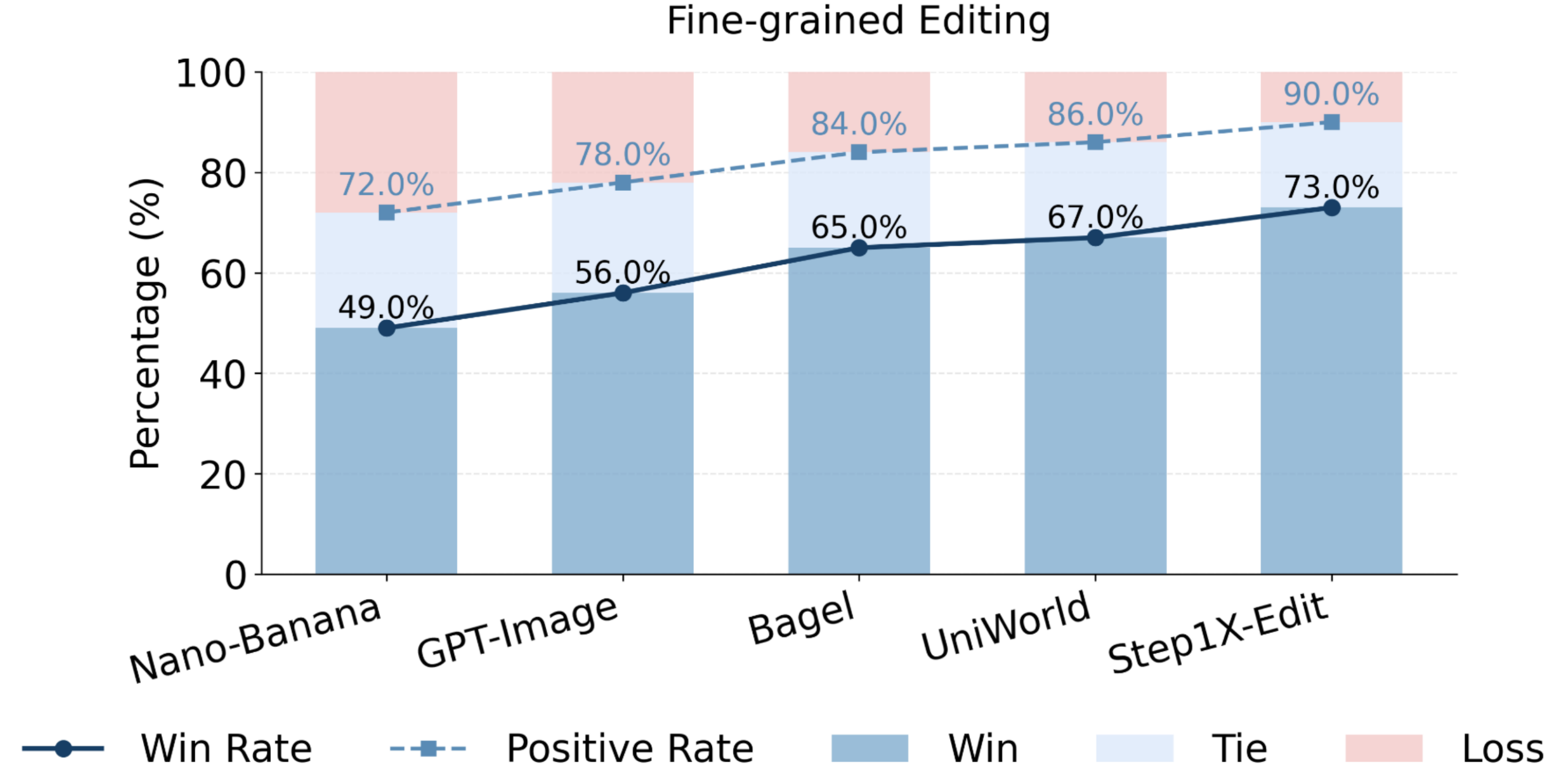}
    \caption{Pairwise human preference study.}
    \label{fig:human_eval}
    \vspace{-0.2cm}
\end{wrapfigure}
\subsubsection{Human Evaluation}
We conduct a human preference study on 200 samples covering diverse editing tasks, with 30 participants performing pairwise comparisons between JarvisEvo and baseline outputs. In each trial, participants see an input image, an instruction, and two edited results in random order, selecting the one that best matches the instruction and aesthetic quality. Each judgment is recorded as a win, loss, or tie for JarvisEvo, from which the win rate (proportion of wins) and positive rate (wins + ties) are computed. We compare several state‑of‑the‑art commercial and open‑source editing models: Nano-Banana~\citep{team2023gemini}, GPT-Image~\citep{hurst2024gpt}, Bagel~\citep{deng2025bagel}, UniWorld-v1~\citep{lin2025uniworld}, Step1X-Edit~\citep{liu2025step1x}.
As depicted in Figure~\ref{fig:human_eval}, JarvisEvo achieved a fine-grained editing win rate of 49\%, surpassing Nano-Banana (28\%) by 21 points. Additionally, it outperformed GPT-Image (22\%), Bagel (16\%), and UniWorld-v1 (14\%) in terms of win rate, indicating its superior performance in the human preference evaluation.

\section{Ablation Study}
\textbf{Analysis of SEPO components.} 
We evaluate five training configurations to quantify each component’s effect: (1) SFT only; (2) SEPO without evaluator optimization; (3) SEPO without selective loss masking (SLM) in the editor loop; (4) SEPO without both evaluator optimization and SLM in the editor loop; (5) Full SEPO with editor–evaluator optimization and SLM.
As shown in rows 3–7 of Table~\ref{tab:ablation} and Figure~\ref{fig:abaltions}, we derive three main insights from the ablation study: 
(1) Without the evaluator optimization loop, the editor's self-evaluation scores gradually rise, the pairwise preference rewards decline steadily, accompanied by a drop in overall editing performance. This indicates that without evaluator calibration, the editor becomes prone to self-deception and reward hacking, generating unreliable reward signals that misdirect the optimization process.
(2) Removing SLM during editor optimization causes self-evaluation scores to rise rapidly and pairwise preference rewards to fall sharply, resulting in unstable training and performance degradation. The underlying reason is self-reward information leakage in the loss computation, which encourages the model to overfit to evaluation cues rather than improving actual editing capability, ultimately leading to optimization collapse.
(3) Lacking both evaluator optimization and SLM causes the system to collapse rapidly during training: the editor’s predicted scores approach near-perfect levels, while editing performance deteriorates substantially. This outcome underscores that both components are essential for maintaining the stability and effectiveness of SEPO.

\begin{table}[!t]
    \centering
    \setlength{\abovecaptionskip}{0.1cm} 
    \setlength{\belowcaptionskip}{-0.3cm}
    \caption{Ablation studies on ArtEdit-Bench-Lr (EN and CN sets).}
    \label{tab:ablation} 
    \scalebox{1}{
        \setlength\tabcolsep{11pt}
        \renewcommand\arraystretch{0.85}
        \begin{tabular}{lccccc}
            \toprule
            \textbf{Configurations} & $\text{L1}_{\times 10^{2}}$ $\downarrow$ & $\text{L2}_{\times 10^{3}}$ $\downarrow$ & $\text{SC} \uparrow$ & $\text{PQ} \uparrow$  & $\text{O} \uparrow$ \\
            \midrule
            \multicolumn{6}{l}{\textcolor{gray!60}{Analysis of SEPO components}} \\ 
            SFT & 9.97 & 19.84 & 8.13 & 8.48 & 8.26\\
            + SEPO w/o Eval. loop & 12.43 & 29.33 & 7.51 & 8.18 & 7.78 \\
            + SEPO w/o SLM & 14.35 & 43.75 & 7.26 & 8.01 & 7.40 \\
            + SEPO w/o SLM and Eval. loop & 17.51 & 59.67 & 6.79 & 7.85 & 7.25 \\
            + \textbf{Full SEPO} & \textbf{8.25} & \textbf{12.84} & \textbf{8.31} & \textbf{8.91} & \textbf{8.54} \\
            \midrule
            \multicolumn{6}{l}{\textcolor{gray!60}{External vs. Intrinsic Rewards}} \\ 
            Gemini as external reward & 10.01 & 20.32 & 7.76 & 8.35 & 8.14 \\
            \textbf{Intrinsic reward} & \textbf{8.25} & \textbf{12.84} & \textbf{8.31} & \textbf{8.91} & \textbf{8.54} \\
            \midrule
            \multicolumn{6}{l}{\textcolor{gray!60}{Effect of different self-reward design}} \\ 
            Absolute point score reward & 8.78 & 13.03 & 8.25 & 8.81 & 8.43 \\
            \textbf{Pairwise preference reward} & \textbf{8.25} & \textbf{12.84} & \textbf{8.31} & \textbf{8.91} & \textbf{8.54} \\
            \midrule
            \multicolumn{6}{l}{\textcolor{gray!60}{Different training stages}} \\ 
            SFT & 9.97 & 19.84 & 8.13 & 8.48 & 8.26\\
            + Full SEPO & 8.25 & 12.84 & 8.31 & 8.91 & 8.54 \\
            + \textbf{Full SEPO + RFT} & \textbf{7.72} & \textbf{11.98} & \textbf{8.53} & \textbf{9.03} & \textbf{8.76} \\

            \bottomrule
        \end{tabular}
    }
    \vspace{-0.1cm} 
\end{table}

\begin{figure}[!t]
    \centering
    \setlength{\abovecaptionskip}{0.05cm} 
    \setlength{\belowcaptionskip}{-0.1cm}
    \includegraphics[width=\linewidth]{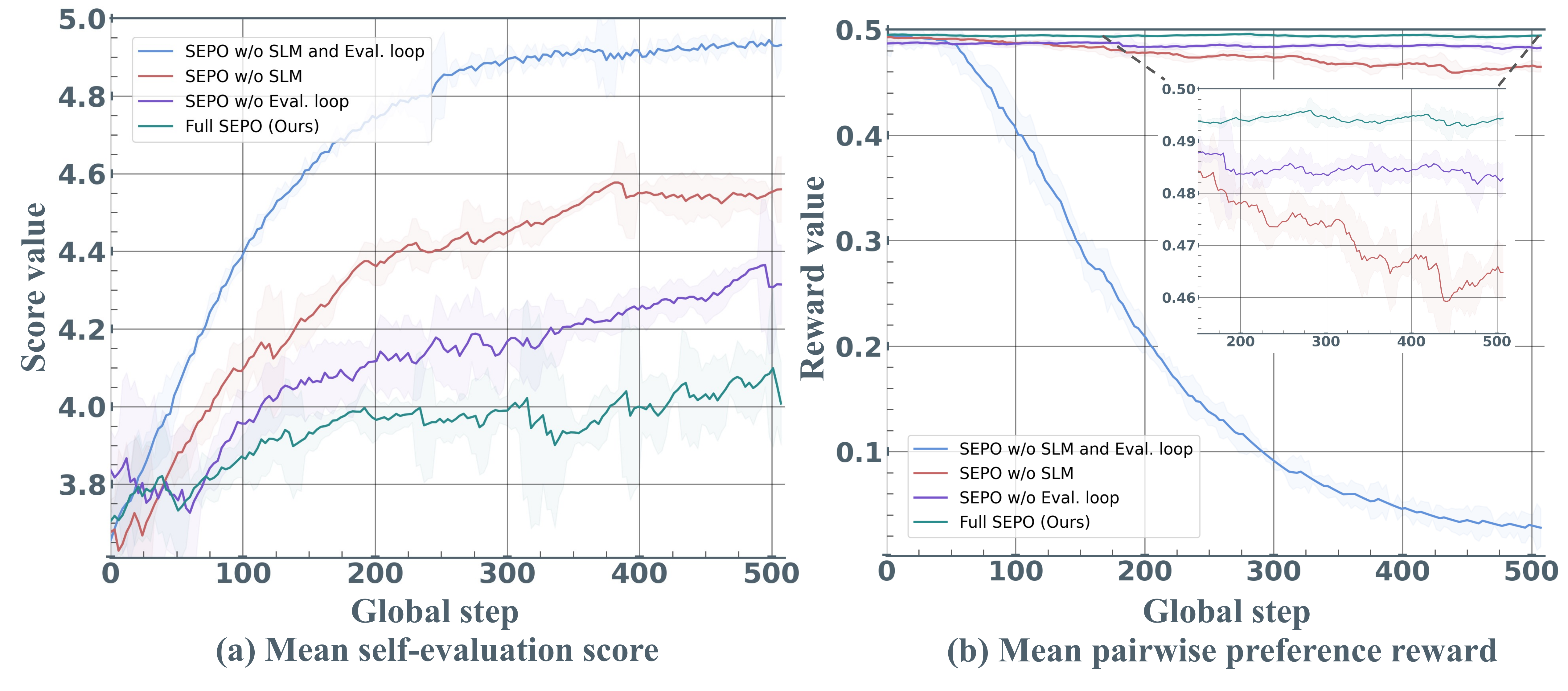} 
    \caption{Trends of self-evaluation scores (left) and pairwise preference rewards (right) across training steps on ArtEdit-Bench-Lr.}
    \label{fig:abaltions}
\end{figure}

\begin{figure}[!t]
    \centering
    \setlength{\abovecaptionskip}{0.02cm} 
    \setlength{\belowcaptionskip}{-0.5cm}
    \includegraphics[width=\linewidth]{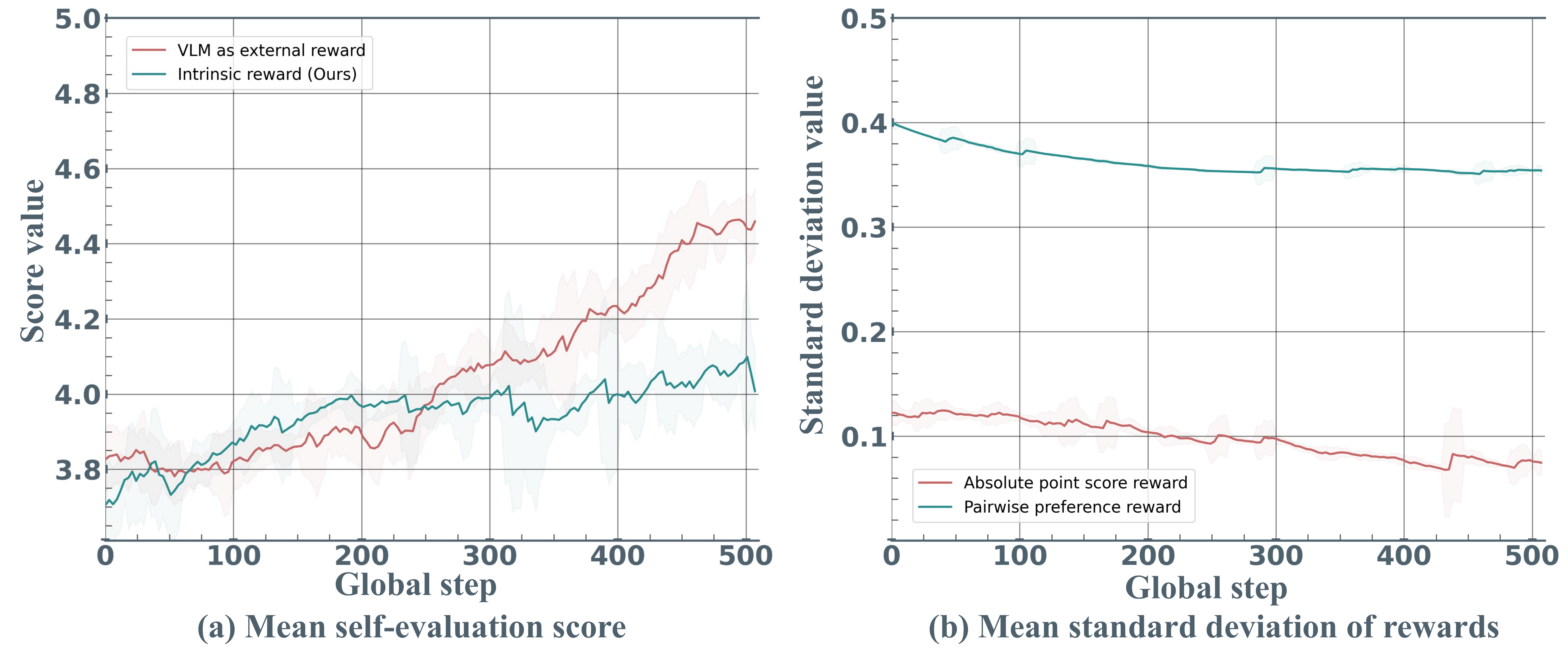} 
    \caption{(a) Training dynamics of self-evaluation scores when using Gemini-2.5-Pro as an external reward model compared with intrinsic rewards. (b) Standard deviation of using absolute point score rewards versus pairwise preference rewards across training.}
    \label{fig:gemini_reward}
\end{figure}

\noindent\textbf{External vs. Intrinsic Reward.} 
We assess the impact of incorporating an external reward model during the SEPO stage. Specifically, we compare two training configurations:
\begin{itemize}
    \item \textbf{VLM as external reward:} The evaluator optimization loop is removed, and Gemini-2.5-Pro~\citep{team2023gemini} is employed as an external reward model to score the images generated by the editor. These scores act as pseudo-labels, $s^{pse}$, used to supervise the model’s predicted scores, $s^{pre}$. 
    The optimization of the editor is guided by several rewards, including format reward, tool accuracy reward, and pairwise preference reward. In this setting, we introduce an additional score alignment reward to supervise the accuracy of the predicted scores. This reward is defined as:
    \begin{equation*}
    R_{sa} = \exp \left( -\frac{1}{2} \left( \frac{|s^{pre} - s^{pse}|}{\sigma} \right)^2 \right) + \epsilon,
    \end{equation*}
    where $\sigma = 0.5$ controls the error tolerance, and $\epsilon$ is a small positive constant to maintain non-zero gradients.

    \item \textbf{Full SEPO framework:} The complete SEPO framework is employed, wherein the editor and evaluator are jointly trained within a unified model.
\end{itemize}

As shown in Table~\ref{tab:ablation} and Figure~\ref{fig:gemini_reward}(a), we observe the following: (1) In the early stages of training, the external VLM (e.g., Gemini-2.5-Pro~\citep{team2023gemini}), acting as a strong AI-based evaluator, provides effective supervision. Under this configuration, the trajectories of the model’s self-predicted scores closely track those observed under the full SEPO setting. (2) In the later stages, however, the model trained with external rewards exhibits clear reward hacking: its self-predicted scores continue to increase, while the degraded editing performance relative to the full SEPO framework. This reward hacking primarily arises from the static nature of the external evaluator. Prior studies have shown that static L(V)LM-as-a-judge setups introduce systematic biases, including cognitive biases (e.g., bandwagon and beauty effects)~\citep{chen2024humans,koo2024benchmarking}, model-inherent biases~\citep{shi2024judging}, self-preference biases (e.g., aesthetic criteria)~\citep{wataoka2024self}, and solution fixation~\citep{li2025curse}, particularly on complex tasks. These biases encourage the editor policy to adapt to the evaluator’s idiosyncratic scoring patterns, leading it to overfit VLM-generated pseudo-reward signals rather than improving true editing quality. As training progresses, the editor exploits the stable scoring behavior of the static reward model, causing a growing misalignment between self-assessed and actual editing performance. In contrast, the co-adaptive design of the full SEPO framework mitigates this issue by dynamically updating the evaluator along with the editor, thereby preserving reward alignment and enhancing generalization.

\noindent\textbf{Impact of pairwise preference reward}
To investigate the effect of using win-rate based pairwise preference reward during the editor optimization loop of SEPO stage, we conduct an ablation study comparing two reward configurations: 
\begin{itemize}
    \item \textbf{Absolute point score as reward:} The editor is optimized using its self-predicted absolute scores as the reward signal, supplemented by format and tool accuracy rewards.
    \item \textbf{Pairwise preference reward (Full SEPO):} The editor is optimized using a pairwise preference reward, which is derived from the win rate of comparisons among multiple generated candidates. This configuration, also supplemented by the other rewards, is the standard setup for our full SEPO framework.
\end{itemize}
The results in Table~\ref{tab:ablation} and Figure~\ref{fig:gemini_reward}(b) highlight three key benefits of using pairwise preference rewards over absolute scores: (1) Amplified reward vriance: by polarizing win rates toward 1 for high-quality edits and 0 for low-quality ones, it enhances separability between reward distributions, yielding more stable and discriminative advantage estimates for policy learning. (2) Reduced Reward Hacking: Since rewards depend on relative comparisons rather than absolute score thresholds, the policy is discouraged from over-optimizing toward superficial scoring patterns, thus mitigating reward hacking. (3) Improved Human Alignment: The pairwise mechanism better reflects human preference judgments between alternatives. This results in reward signals that capture nuanced editorial preferences more accurately, ultimately improving both user alignment and overall editing quality.

\noindent\textbf{Different training stages.} Rows 15–17 of Table~\ref{tab:ablation} demonstrate the effectiveness of various stages in our training pipeline. SEPO and RFT consistently outperform the SFT model across all evaluation metrics.


\section{Conclusions}
\label{sec:conclusions}
We introduce JarvisEvo, a co-evolutionary editing agent that integrates the roles of editor and evaluator within a single model. JarvisEvo contributes in three significant ways: (1) it incorporates an interleaved multimodal Chain-of-Thought (iMCoT) reasoning mechanism, which improves both instruction following and the quality of editing; (2) it features a synergistic editor–evaluator policy optimization (SEPO) framework that facilitates self-improvement without the need for external rewards, thus reducing the risk of reward hacking; and (3) It supports both global and local professional editing through its integration with Adobe Lightroom, enabling precise and detailed aesthetic retouching.

\section{Limitation and Broader Impacts}
\label{app:limitation}
While JarvisEvo demonstrates the feasibility and effectiveness of the synergistic editor-evaluator co-evolution paradigm within a unified model—yielding promising results in editing capabilities, self-evaluation, self-reflection, and multimodal reasoning—several limitations remain that warrant further exploration.
First, although our current implementation focuses on photo editing, the core philosophy of synergistic self-improvement is task-agnostic. The paradigm of coupling a generator (editor) with an internal critic (evaluator) to facilitate mutual enhancement holds significant potential for a broader range of Large Language Model (LLM) tasks. We believe this approach could be effectively adapted to other complex domains, such as mathematical reasoning, code generation, and long-horizon planning, where self-correction and iterative refinement are crucial. Future work could investigate the generalizability of this co-evolutionary framework to these diverse applications.
Second, our current interleaved Multimodal Chain-of-Thought (iMCoT) reasoning is limited to a maximum of four editing steps. Extending this to longer-horizon tasks (e.g., exceeding 10 steps) remains unexplored due to the significant challenges in constructing high-quality, multi-step training data and the prohibitive computational resources required for such extended reasoning sequences.

\section{Contributors}
\begin{itemize}
    \item \textbf{Project leaders:} Qinglin Lu, Chunyu Wang
    \item \textbf{Core contributors:} Yunlong Lin, Linqing Wang, Kunjie Lin, Zixu Lin
    \item \textbf{Corresponding author:} Xinghao Ding
    \item \textbf{Contributors:} Kaixiong Gong, Wenbo Li, Bin Lin, Zhenxi Li, Shiyi Zhang, Yuyang Peng, Wenxun Dai
\end{itemize}





\clearpage

\clearpage
    \begin{longtable}{>{\RaggedRight}p{0.25\textwidth} >{\RaggedRight}p{0.4\textwidth} >{\RaggedRight}p{0.2\textwidth} >{\RaggedRight\arraybackslash}p{0.07\textwidth}}
    \caption{Lightroom tools with functional description, numerical range, and parameter type.}\label{tab:tools}\\
    \toprule
    \textbf{Tool Name} & \textbf{Functional Description} & \textbf{Numerical Range} & \textbf{Type} \\
    \midrule
    \endfirsthead
    \caption[]{Lightroom tools with functional description, numerical range, and parameter type. (Continued)}\\
    \toprule
    \textbf{Tool Name} & \textbf{Functional Description} & \textbf{Numerical Range} & \textbf{Type} \\
    \midrule
    \endhead
    \midrule
    \multicolumn{4}{r}{\textit{Continued on next page}} \\
    \endfoot
    \bottomrule
    \endlastfoot
    
    \multicolumn{4}{c}{\param{\textbf{Basic Adjustments}}} \\
    WhiteBalance & Overall color temperature & As Shot, Auto, Custom & Str. \\
    Temperature & Blue-yellow balance & 2000-10000 Kelvin & Num. \\
    Tint & Green-magenta balance & -150 to +150 & Num. \\
    Exposure2012 & Overall brightness & -5.0 to +5.0 stops & Num. \\
    Contrast2012 & Difference between light/dark areas & -100 to +100 & Num. \\
    Highlights2012 & Adjusts bright areas & -100 to +100 & Num. \\
    Shadows2012 & Adjusts dark areas & -100 to +100 & Num. \\
    Whites2012 & Fine-tunes brightest parts & -100 to +100 & Num. \\
    Blacks2012 & Fine-tunes darkest parts & -100 to +100 & Num. \\
    Texture & Enhances/smooths medium textures & -100 to +100 & Num. \\
    Clarity2012 & Enhances/reduces local mid-tone contrast & -100 to +100 & Num. \\
    Dehaze & Reduces/adds atmospheric haze & -100 to +100 & Num. \\
    Vibrance & Saturation of less-saturated colors & -100 to +100 & Num. \\
    Saturation & Overall color intensity & -100 to +100 & Num. \\
    IncrementalTemperature & Relative temperature adjustment & -100 to +100 & Num. \\
    IncrementalTint & Relative tint adjustment & -100 to +100 & Num. \\
    \midrule
    
    \multicolumn{4}{c}{\param{\textbf{Tone Curve}}} \\
    ToneCurveName2012 & Predefined curve shape & Linear, Custom & Str. \\
    ToneCurvePV2012 & Custom RGB tone curve points & x,y: 0-255 & Dict. \\
    ToneCurvePV2012Red & Custom Red channel tone curve points & x,y: 0-255 & Dict. \\
    ToneCurvePV2012Green & Custom Green channel tone curve points & x,y: 0-255 & Dict. \\
    ToneCurvePV2012Blue & Custom Blue channel tone curve points & x,y: 0-255 & Dict. \\
    ParametricShadows & Adjusts shadow tonal regions & -100 to +100 & Num. \\
    ParametricDarks & Adjusts dark tonal regions & -100 to +100 & Num. \\
    ParametricLights & Adjusts light tonal regions & -100 to +100 & Num. \\
    ParametricHighlights & Adjusts highlight tonal regions & -100 to +100 & Num. \\
    ParametricShadowSplit & Boundary: shadows/darks & 10-50 & Num. \\
    ParametricMidtoneSplit & Boundary: darks/lights & 25-75 & Num. \\
    ParametricHighlightSplit & Boundary: lights/highlights & 50-90 & Num. \\
    \midrule
    
    \multicolumn{4}{c}{\param{\textbf{Detail}}} \\
    Sharpness & Enhances edge definition & 0-150 & Num. \\
    SharpenRadius & Width of sharpening effect & 0.5-3.0 & Num. \\
    SharpenDetail & Amount of sharpening for details & 0-100 & Num. \\
    SharpenEdgeMasking & Masks sharpening to edges & 0-100 & Num. \\
    LuminanceSmoothing & Reduces luminance noise & 0-100 & Num. \\
    ColorNoiseReduction & Reduces color noise & 0-100 & Num. \\
    ColorNoiseReduce.Detail & Fine-tunes color noise reduction & 0-100 & Num. \\
    ColorNoiseReduce.Smooth & Smoothness of color noise reduction & 0-100 & Num. \\
    \midrule
    
    \multicolumn{4}{c}{\param{\textbf{HSL/Color (per color: Red, Orange, Yellow, Green, Aqua, Blue, Purple, Magenta)}}} \\
    HueAdjustment & Shifts hue of specific color & -100 to +100 & Num. \\
    SaturationAdjustment & Adjusts saturation of specific color & -100 to +100 & Num. \\
    LuminanceAdjustment & Adjusts brightness of specific color & -100 to +100 & Num. \\
    \midrule
    
    \multicolumn{4}{c}{\param{\textbf{Color Grading}}} \\
    SplitToningShadowHue & Hue for shadows in split toning & 0-359 & Num. \\
    SplitToningHighlightHue & Hue for highlights in split toning & 0-359 & Num. \\
    SplitToningShadowSat. & Saturation for shadows & 0-100 & Num. \\
    SplitToningHighlightSat. & Saturation for highlights & 0-100 & Num. \\
    SplitToningBalance & Balance between shadow/highlight toning & -100 to +100 & Num. \\
    ColorGradeMidtoneHue & Midtone hue for color grading & 0-359 & Num. \\
    ColorGradeMidtoneSat & Midtone saturation for color grading & 0-100 & Num. \\
    ColorGradeMidtoneLum & Midtone luminance for color grading & 0-100 & Num. \\
    ColorGradeShadowLum & Luminance for shadows & 0-100 & Num. \\
    ColorGradeHighlightLum & Luminance for highlights & 0-100 & Num. \\
    ColorGradeBlending & Blending of color grading effect & 0-100 & Num. \\
    ColorGradeGlobalHue & Global hue adjustment & 0-359 & Num. \\
    ColorGradeGlobalSat & Global saturation adjustment & 0-100 & Num. \\
    ColorGradeGlobalLum & Global luminance adjustment & 0-100 & Num. \\
    \midrule
    
    \multicolumn{4}{c}{\param{\textbf{Effects}}} \\
    PostCropVignetteAmount & Darkens/lightens image corners & -100 to +100 & Num. \\
    GrainAmount & Adds film grain effect & 0-100 & Num. \\
    ShadowTint & Adjusts color tint in shadows & -100 to +100 & Num. \\
    \midrule
    
    \multicolumn{4}{c}{\param{\textbf{Camera Calibration (for Red, Green, Blue primary channels)}}} \\
    <PrimaryColor>Hue & Shifts primary color's hue & -100 to +100 & Num. \\
    <PrimaryColor>Saturation & Adjusts primary color's saturation & -100 to +100 & Num. \\
    \midrule
    
    \multicolumn{4}{c}{\param{\textbf{Lens Blur (Overall: Dict.)}}} \\
    LensBlur.Active & Enables/disables lens blur effect & -- & Bool. \\
    LensBlur.BlurAmount & Strength of blur effect & 0-100 & Num. \\
    LensBlur.FocalRange & Defines focal plane & "x1 y1 x2 y2" & Str. \\
    LensBlur.BokehShape & Bokeh shape identifier & default 0 & Num. \\
    LensBlur.BokehShape & Definition of bokeh shape edges & 0-100 & Num. \\
    LensBlur.Highlights & Brightness threshold for bokeh & 0-100 & Num. \\
    LensBlur.HighlightsBoost & Enhances out-of-focus highlights & 0-100 & Num. \\
    LensBlur.CatEyeAmount & Simulates cat's eye bokeh effect & 0-100 & Num. \\
    LensBlur.CatEyeScale & Size of cat's eye effect & 0-100 & Num. \\
    \midrule
    
    \multicolumn{4}{c}{\param{\textbf{Advanced Color Grading (PointColors - each point is a Dict.)}}} \\
    SrcHue & Source hue for adjustment & 0-6.28 rad & Num. \\
    SrcSat & Source saturation for adjustment & 0-1.0 & Num. \\
    SrcLum & Source luminance for adjustment & 0-1.0 & Num. \\
    HueShift & Hue shift amount & -1 to +1 & Num. \\
    SatScale & Saturation scale & -1 to +1 & Num. \\
    LumScale & Luminance scale & -1 to +1 & Num. \\
    RangeAmount & Effect application amount & 0-1.0 & Num. \\
    HueRange & Falloff for hue adjustment & 0-1.0 & Dict. \\
    SatRange & Falloff for saturation adjustment & 0-1.0 & Dict. \\
    LumRange & Falloff for luminance adjustment & 0-1.0 & Dict. \\
    \midrule
    
    \multicolumn{4}{c}{\param{\textbf{Look (Overall: Dict.)}}} \\
    Look.Name & Name of the look preset & -- & Str. \\
    Look.Amount & Intensity of the look effect & 0.0-1.0 & Num. \\
    Look.Parameters & Dictionary of specific adjustments & -- & Dict. \\
    \ \ \textit{ (e.g., ProcessVersion, ToneCurvePV2012, etc.)} & & & \\
    \midrule
    
    \multicolumn{4}{c}{\param{\textbf{Localized Mask Adjustments (MaskGroupBasedCorrections - Array of Dicts.)}}} \\
    \textit{Per Correction Group:} \\
    CorrectionAmount & Amount for the correction group & 0-1 & Num. \\
    CorrectionActive & Activates the correction group & -- & Bool. \\
    CorrectionName & Name for the correction group & -- & Str. \\
    LocalExposure2012 & Local exposure adjustment & -1 to +1 & Num. \\
    LocalContrast2012 & Local contrast adjustment & -1 to +1 & Num. \\
    LocalHighlights2012 & Local highlights adjustment & -1 to +1 & Num. \\
    LocalShadows2012 & Local shadows adjustment & -1 to +1 & Num. \\
    LocalWhites2012 & Local whites adjustment & -1 to +1 & Num. \\
    LocalBlacks2012 & Local blacks adjustment & -1 to +1 & Num. \\
    LocalClarity / LocalClarity2012 & Local clarity adjustment & -1 to +1 & Num. \\
    LocalDehaze & Local dehaze adjustment & -1 to +1 & Num. \\
    LocalTexture & Local texture adjustment & -1 to +1 & Num. \\
    LocalHue & Local hue adjustment & -1 to +1 & Num. \\
    LocalSaturation & Local saturation adjustment & -1 to +1 & Num. \\
    LocalCurveRefineSat. & Local saturation curve refinement & 0-100 & Num. \\
    LocalToningHue & Local toning hue & 0-359 & Num. \\
    LocalToningSaturation & Local toning saturation & -1 to +1 & Num. \\
    LocalTemperature & Local temperature adjustment & -1 to +1 & Num. \\
    LocalTint & Local tint adjustment & -1 to +1 & Num. \\
    LocalLuminanceNoise & Local luminance noise reduction & -1 to +1 & Num. \\
    LocalMoire & Local moire reduction & -1 to +1 & Num. \\
    LocalDefringe & Local defringe adjustment & -1 to +1 & Num. \\
    LocalGrain & Local grain adjustment & -1 to +1 & Num. \\
    LocalSharpness & Local sharpness adjustment & -1 to +1 & Num. \\
    <Channel>Curve & Local tone curve for channels & "x,y" points & Dict. \\
    LocalPointColors & Local specific color adjustments & -- & Dict. \\
    CorrectionMasks & Array of mask definitions & -- & Array \\
    \ \ \textit{Per Mask in CorrectionMasks:} \\
    \ \ What & Mask type & e.g., "Mask/Image" & Str. \\
    \ \ MaskActive & Activates this specific mask & -- & Bool. \\
    \ \ MaskName & Name of the mask & e.g., "Subject" & Str. \\
    \ \ MaskBlendMode & Mask blending & 0=Add, 1=Intersect & Num. \\
    \ \ MaskInverted & Inverts the mask area & -- & Bool. \\
    \ \ MaskValue & Mask opacity & 0.0-1.0 & Num. \\
    \ \ MaskSubType & AI Mask subtype / Object type & -- & Num. \\
    \ \ ReferencePoint & Center point for AI masks & "x y" & Str. \\
    \ \ Gesture & Polygon points for object/region mask & -- & Array \\
    \ \ Top/Left/Bottom/Right & Coordinates for radial gradient & 0-1 & Num. \\
    \ \ Angle & Rotation angle for radial gradient & 0-360 & Num. \\
    \ \ Midpoint & Center point of radial gradient & 0-100 & Num. \\
    \ \ Feather & Edge feathering for radial gradient & 0-100 & Num. \\
    \ \ Flipped & Flips radial gradient direction & -- & Bool. \\
    \ \ MaskSubCategoryID & Category ID for person parts mask & -- & Num. \\
\end{longtable}
\clearpage

\clearpage

\bibliography{colm2024_conference}
\bibliographystyle{colm2024_conference}
\end{document}